%% file: main.tex
\pdfoutput=1

\documentclass[11pt]{article}

\usepackage[]{EMNLP2022}

\usepackage{times}
\usepackage{latexsym}

\usepackage[T1]{fontenc}

\usepackage[utf8]{inputenc}

\usepackage{microtype}

\usepackage{inconsolata}

\input{tools}

\title{Towards Robust Numerical Question Answering: Diagnosing~Numerical~Capabilities~of~NLP~Systems}

\makeatletter
\newcommand{\printfnsymbol}[1]{%
  \textsuperscript{\@fnsymbol{#1}}%
}
\makeatother

\author{
Jialiang Xu\textsuperscript{\rm 1}\thanks{\indent The contributions by Jialiang Xu and Xinyi He have been conducted and completed during their internships at Microsoft Research Asia, Beijing, China.}, 
Mengyu Zhou\textsuperscript{\rm 2}\thanks{\indent Corresponding author.}, 
Xinyi He\textsuperscript{\rm 3}\printfnsymbol{1}, 
Shi Han\textsuperscript{\rm 2}, 
Dongmei Zhang\textsuperscript{\rm 2} \\
\textsuperscript{\rm 1} University of Illinois at Urbana-Champaign \\
\textsuperscript{\rm 2} Microsoft Research 
\textsuperscript{\rm 3} Xi'an Jiaotong University \\ 
\texttt{\href{mailto:jx17@illinois.edu}{jx17@illinois.edu}}, \texttt{\href{mailto:hxyhxy@stu.xjtu.edu.cn}{hxyhxy@stu.xjtu.edu.cn}},\\ \texttt{\{\href{mailto:mezho@microsoft.com}{mezho}, \href{mailto:shihan@microsoft.com}{shihan}, \href{mailto:dongmeiz@microsoft.com}{dongmeiz}\}@microsoft.com}}

\begin{document}


\maketitle
\begin{abstract}
\input{body/0-Abstract}
\end{abstract}

\section{Introduction}
\input{body/1-Introduction}

\section{Related Work}
\input{body/2-RelatedWork}

\section{Preliminaries}
\label{sec:Preliminaries}
\input{body/3-Preliminaries}

\section{DNC Framework}
\input{body/4-Methodology}

\input{body/table_main_exp}
\section{Experiments}
\label{sec:exp}
\input{body/5-Experiments}

\section{Guidelines and Open Directions}
\label{sec:emp}
\input{body/6-Potential_Directions}


\section{Conclusion}
\input{body/7-Conclusion}

\section*{Limitations}
Our pipeline has limitations in the following two aspects that we plan to address in the future:

\textbf{Dependency on ground truth equation}. Currently, three of the eight DNC perturbations have strong dependency on the ground truth solving equation, which is missing in datasets such as DROP. We hope to utilize semi-supervised approaches in the future to enlarge the coverage of the DNC perturbations.

\textbf{Perturbing scalability}. Currently our filters cover only a portion of the whole dataset due to DNC filtering and perturbing questions based on manual rules and templates. we hope to develop more automatic filtering and perturbing in the future. Also, DNC can only apply perturbations to numbers provided by the problem, which limits its diagnosing power in questions where an unspecified number is used, \eg, when numerical commonsense knowledge is involved.

\section*{Ethical Statements}
The model implementation and datasets utilized in this paper are based on publication and open-source repositories. Licenses protocols are followed in the process of our experiments. No new datasets or NLP applications are presented in this paper and no violation of privacy or usage of demographic information was involved in our process of interacting with the datasets. Our experiments do not involve lots of compute time/power as reported in the paper.
\clearpage


\input{main.bbl}
\clearpage
\appendix
\input{body/Appendix}

\end{document}

%% file: tools.tex
\usepackage{amsmath}
\usepackage{multirow}
\usepackage{multicol}
\usepackage{colortbl}
\usepackage{xspace}
\usepackage{graphicx}
\usepackage{booktabs}
\usepackage{algorithm2e}
\usepackage{circledsteps}
\usepackage{amsmath}
\usepackage{amssymb}

\newcommand{\original}[1]{\textcolor{brown}{#1}}
\newcommand{\perturbed}[1]{\textcolor{teal}{#1}}

\newcommand{\correcteq}{$\textcolor{teal}{\checkmark}$}
\newcommand{\wrongeq}{$\textcolor{purple}{\times}$}

\newcommand{\perfdrop}[2]{#1\% absolute drop and #2\% relative drop}
\newcommand{\perfimprove}[2]{#1\% absolute improvement and #2\% relative improvement}

\newcommand{\reffig}[1]{Figure~\ref{#1}}
\newcommand{\refsec}[1]{\S\ref{#1}} 
\newcommand{\reftab}[1]{Table~\ref{#1}}

\newcommand{\refapp}[1]{Appendix~\ref{#1}}

\def\eg{\textit{e.g.}\xspace}
\def\Eg{\textit{E.g.}\xspace}

\def\etc{\textit{etc.}\xspace}
\def\ie{\textit{i.e.}\xspace}
\def\Ie{\textit{I.e.}\xspace}
\def\vs{\textit{vs.}\xspace}
\def\wrt{\textit{w.r.t.}\xspace}

%% file: body/0-Abstract.tex
Numerical Question Answering is the task of answering questions that require numerical capabilities. Previous works introduce general adversarial attacks to Numerical Question Answering, while not systematically exploring numerical capabilities specific to the topic. In this paper, we propose to conduct numerical capability diagnosis on a series of Numerical Question Answering systems and datasets. A series of numerical capabilities are highlighted, and corresponding dataset perturbations are designed. Empirical results indicate that existing systems are severely challenged by these perturbations. \Eg, Graph2Tree experienced a 53.83\% absolute accuracy drop against the ``Extra'' perturbation on ASDiv-a, and BART experienced 13.80\% accuracy drop against the ``Language'' perturbation on the numerical subset of DROP. As a counteracting approach, we also investigate the effectiveness of applying perturbations as data augmentation to relieve systems' lack of robust numerical capabilities. With experiment analysis and empirical studies, it is demonstrated that Numerical Question Answering with robust numerical capabilities is still to a large extent an open question. We discuss future directions of Numerical Question Answering and summarize guidelines on future dataset collection and system design.



%% file: body/1-Introduction.tex
\begin{figure*}[!ht]
    \centering
    \includegraphics[width=\textwidth]{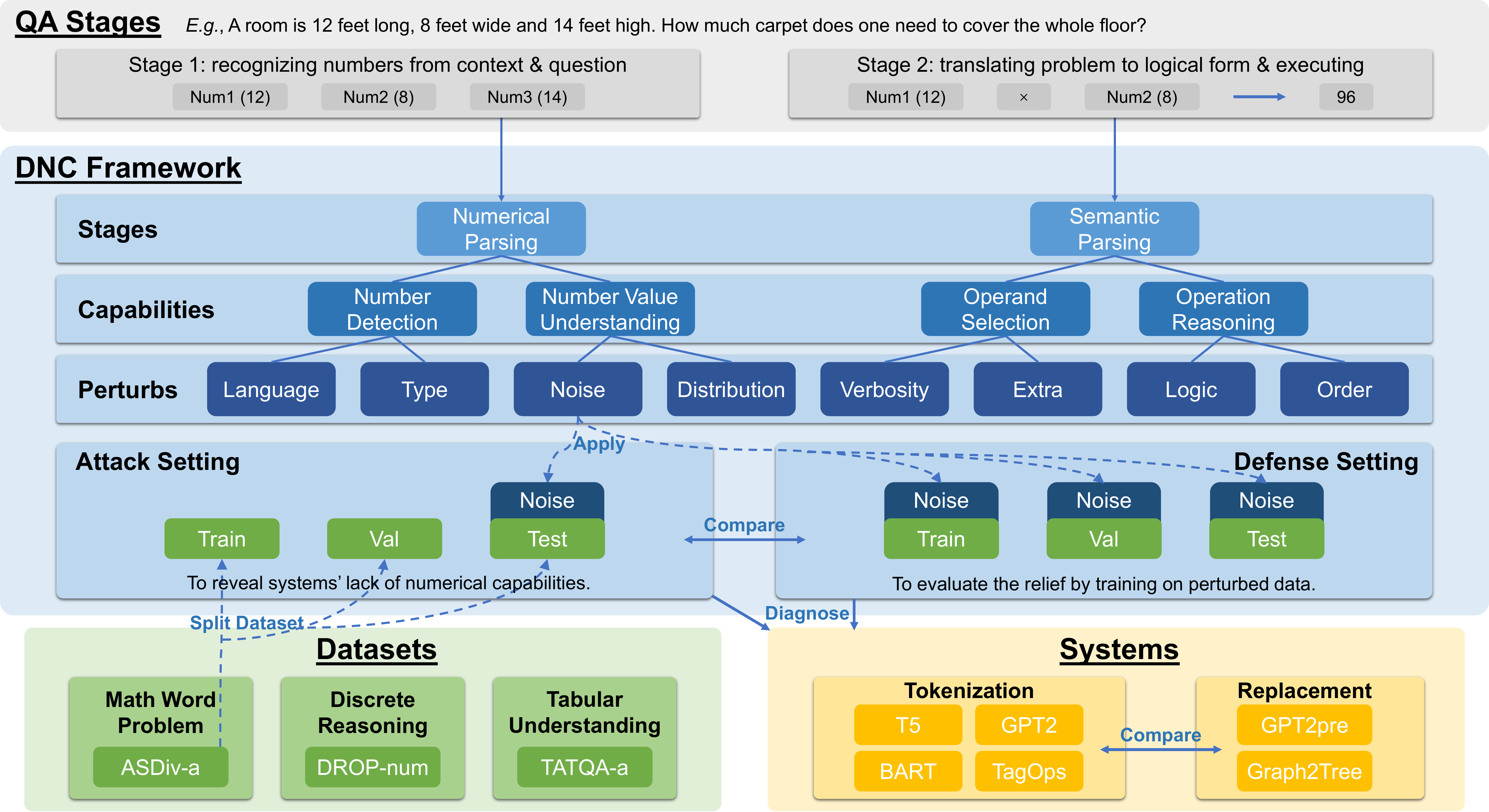}
    \caption{Overview of DNC Framework. The process of Numerical QA solving is divided into two logical stages. Four capabilities are required to complete the stages, each maps to two perturbations. Perturbations can be applied to appropriate train / validation / test splits of Numerical QA datasets under Attack or Defense Setting. Models of the NLP systems are trained and then evaluated on the perturbed datasets as a diagnosis of their numerical capabilities.}
    \label{fig:overview}
\end{figure*}

Numeracy is an essential part for real-world NLP applications \citep{Sundaram2022WhyAN, thawani-etal-2021-representing,  sundararaman-etal-2020-methods, spithourakis-riedel-2018-numeracy}. \textbf{Numerical QA} (Question Answering) is one representative group of such number-dependent NLP tasks. \Eg, Math Word Problem Solving \citep{8703135, miao-etal-2020-diverse, koncel-kedziorski-etal-2016-mawps}, Discrete Reasoning \citep{dua-etal-2019-drop, hu-etal-2019-multi, al-negheimish-etal-2021-discrete}, Tabular Question Answering \citep{zhongSeq2SQL2017, 2019TabFactA, zhu-etal-2021-tat, chen-etal-2021-finqa}. These Numerical QA tasks require NLP systems to arrive at a numerical answer from the numbers in the question and context. 
By studying how existing NLP systems perform in these Numerical QA tasks, we could take a glimpse at what capabilities are required for building NLP systems in the future.


In an ad-hoc manner, a line of work revealed robustness issues of handling Numerical QA in existing NLP systems. Through adversarial attacks with designed dataset perturbations, number-related limitations were exposed: \Eg, utilizing spurious correlation in datasets \citep{patel-etal-2021-nlp, kumar-etal-2021-adversarial-examples, al-negheimish-etal-2021-numerical, pi-etal-2022-towards}, incorrectly representing numbers \citep{DBLP:journals/corr/abs-2102-13019, kim-etal-2021-seen} and failing to extrapolate \citep{kim-etal-2021-seen, pal-baral-2021-investigating-numeracy}. This line of work inspires us to ask following questions: 
1) What is the overall landscape of \textit{robustness issues of numerical capabilities} in existing NLP systems? Can we find a more systematic way to investigate the number-related limitations? 
2) How to \textit{diagnose} each numerical capability and \textit{evaluate} the severity of it not being captured in a system? Can we further develop new adversarial \textit{perturbation} methods on Numerical QA for diagnosis and evaluation? 
3) How to \textit{address} the numerical robustness issues? How do \textit{existing solutions} work and what are possible \textit{future directions}?



To answer the above questions, in this paper we propose the \textbf{DNC} (\textbf{D}iagnosing \textbf{N}umerical \textbf{C}apbilities) \textbf{framework}\footnote{Our code and datasets used are available in the link: https://github.com/microsoft/NumberDiagnosis} as shown in \reffig{fig:overview}. 

Most existing Numerical QA systems (see \refsec{sec:NQA-related}) take a two-stage approach to extract and manipulate numbers. As shown in the \underline{\textbf{\texttt{QA Stages}}} part of \reffig{fig:overview}, 
systems usually first recognize numbers in the context and question and treat them as candidate operands. Then, with the understanding of the question semantics, they select corresponding operands, and explicitly generate logical forms or implicitly execute operations to get the final result.

The above two stages correspond to the two groups of numerical capabilities (see \refsec{Numerical Capabilities}) covered by our \underline{\textbf{\texttt{DNC Framework}}} (as shown in \reffig{fig:overview}). In Stage 1, we focus on a system's capabilities to recognize different forms of numbers (``Number Detection''), and to parse and represent number values correctly (``Number Value Understanding''). In Stage 2, we focus on the capabilities to correctly choose operands (``Operand Selection'') and operations (``Operation Reasoning'') by understanding context and question. 
For each of these four capabilities, two perturbations (see \refsec{Perturbations}) are proposed by us to diagnose the capability. Each perturbation is designed to be trivial to humans and thus cannot easily fool humans, but it could bring down existing NLP systems (under the ``Attack'' setting), and therefore expose the robustness issue of lacking its corresponding capability.
By applying the above diagnosis to various NLP \underline{\textbf{\texttt{Systems}}} and Numerical QA \underline{\textbf{\texttt{Datasets}}} (as shown in \reffig{fig:overview}), in \refsec{sec:exp} we find that existing systems experience significant performance drops, which verifies their lack of robust numerical capabilities. \Eg, Graph2Tree experienced a 53.83\% absolute accuracy drop against the ``Extra'' perturbation on ASDiv-a, and BART experienced 13.80\% accuracy drop against the ``Language'' perturbation on the numerical subset of DROP.

From another point of view, the perturbations are also applicable for data augmentation. Under the ``Defense'' setting (see \refsec{Perturbing Settings}), perturbations are applied to all splits of the dataset. A system's performance of the same perturbation under both ``Attack'' and ``Defense'' settings are compared (in \refsec{sec:exp_analysis}) to show if the corresponding robustness issue could be relieved by augmenting the training data. Empirical results indicate that despite the recovery in most cases, the performance still fall lower than the original level.

Finally, based on the ``Attack'' and ``Defense'' results in \refsec{sec:exp} and additional experiments, in \refsec{sec:emp} we compare some existing design choices in Numerical QA, such as: Is it better to generate logical forms (and then execute the program/expression) or predict answers directly in an end-to-end way? Shall we break numbers into subword tokens or substitute them with a placeholder that can be later re-substituted? 
We also discuss the open questions and future directions on the robust numerical capabilities of NLP systems, including recent relevant development such as neural program execution and numerical data synthesizing.


In summary, our major contributions are:
\begin{itemize}
    \item The DNC framework is proposed by us to systematically diagnose the robustness of NLP systems on numerical capabilities. A series of number-related perturbation methods are designed for the capabilities.
    \item Comprehensive diagnosing experiments on adversarial attacks and data augmentations are conducted by us on five systems over three Numerical QA tasks. We show the overall picture of numerical robustness issues of the systems, and the partial effectiveness of our simple defense mechanism.
    \item Based on experiments and previous work, we provide guidelines for existing numerically-robust NLP system designs and discussions for future directions on robust Numerical QA.
\end{itemize}

%% file: body/2-RelatedWork.tex
\subsection{Numerical Question Answering}
\label{sec:NQA-related}
Previous work has proposed Numerical QA datasets and systems. In this paper we consider as examples the domains of Math Word Problem, Discrete Reasoning and Tabular QA.

\textbf{Math Word Problem} \cite{kushman-etal-2014-learning, upadhyay-chang-2017-annotating, miao-etal-2020-diverse,  qin-etal-2020-semantically, Lan_Wang_Zhang_Lan_Dai_Wang_Zhang_Lim_2022} concerns arithmetic questions collected from lower-grade elementary school coursework. Neural network are employed with different architectures such as Seq2Seq \citep{wang-etal-2017-deep, chiang-chen-2019-semantically}, Seq2Tree \citep{ijcai2019-736, DBLP:journals/corr/abs-2107-13435} and Graph2Tree \citep{zhang-etal-2020-graph-tree, shen-jin-2020-solving}. Recently, large end-to-end pretrained language models \citep{Chowdhery2022PaLMSL, Pi2022ReasoningLP} have also been showing impressive results in Math Word Problem.

\textbf{Discrete Reasoning} \cite{dua-etal-2019-drop, al-negheimish-etal-2021-discrete, hu-etal-2019-multi} concerns questions requiring logistic and arithmetic operations on real-world paragraphs. Discrete Reasoning Systems are mainly based on Graph Attention Networks \citep{chen-etal-2020-question} or the Transformer architecture \citep{ran-etal-2019-numnet}.

\textbf{Tabular QA and Semantic Parsing} \cite{zhu-etal-2021-tat, chen-etal-2021-finqa, zhongSeq2SQL2017, pasupat-liang-2015-compositional} concerns question answering in the domain of tabular data, which often involves a large amount of numbers and requires arithmetic aggregations to arrive at the final answer. Tabular QA systems \citep{dong2022table, liu2022tapex, iida-etal-2021-tabbie, herzig-etal-2020-tapas, yin-etal-2020-tabert} are mainly based on Pretrained Language Models with Transformer backbones. Tabular QA systems mainly aim at converting natural language utterance into executable expressions such as commands in SQL language.

\subsection{Numeracy Limitations in NLP Systems}
Efforts have been dedicated to reveal numeracy limitations in NLP systems. \citep{patel-etal-2021-nlp, kumar-etal-2021-adversarial-examples, al-negheimish-etal-2021-numerical, pi-etal-2022-towards, DBLP:journals/corr/abs-2102-13019, kim-etal-2021-seen, pal-baral-2021-investigating-numeracy}. However, previous work mainly focused on borrowing adversarial attack methods from general QA such as re-ordering sentences \citep{patel-etal-2021-nlp, al-negheimish-etal-2021-numerical, kumar-etal-2021-adversarial-examples}, substituting synonyms  \citep{kumar-etal-2021-adversarial-examples, pi-etal-2022-towards}, or adding irrelevant information \citep{patel-etal-2021-nlp, pi-etal-2022-towards}, while having limited exploration into capabilities specific to Numerical QA problems such as understanding different number values, recognizing different number surface forms or selecting related numbers. 





%% file: body/3-Preliminaries.tex

A \textit{Numerical Question Answering} problem is defined to consist of a problem prompt (question) $\mathcal{P}$ and a problem body (context) $\mathcal{B}$. Depending on the task type, the problem body takes the form of either a paragraph or a mixture of free-form text paragraphs and structured data such as tables. Let $\mathcal{V}$ be the vocabulary of the textual words, $\mathcal{Q}$ be the set of the numerical values in $\mathcal{P} \cup \mathcal{B}$, and $\mathcal{Q^+}$ be the numerical values that can be arithmetically computed with $\mathcal{Q}$, then the problem prompt and body can be formulated as $\mathcal{P} = \bigoplus_{i} p_i, ~p_i \in \mathcal{V} \cup \mathcal{Q}$ and $\mathcal{B} = 
    \begin{cases} 
        \bigoplus_{j} b_j \\
        \bigoplus_{i,j} \tau_{i,j} \oplus \bigoplus_{k} b_k
    \end{cases}, ~\tau_{i,j}, b_k \in \mathcal{V} \cup \mathcal{Q}$.
Here $\oplus$ denotes the concatenation operation, $p_{\cdot}$ and $b_{\cdot}$ are prompt and body textual words, and $\tau_{\cdot,\cdot}$ are the body tabular cells.

The target output $\mathcal{T}$ of the problem is either a numerical value $\mathcal{T}_{ans}$ that is an element in $\mathcal{Q^+}$ or a mathematical expression $\mathcal{T}_{eq}$ that consists of elements in the concerned numerical values $\mathcal{Q}$ and the simple operators $\mathcal{O} = \{+, -, \times, \div\}$.
\Ie $\mathcal{T} = 
    \begin{cases}
        \mathcal{T}_{ans}: q \in \mathcal{Q^+} \\
        \mathcal{T}_{eq}: \bigoplus_{i} t_i, ~t_i \in \mathcal{Q} \cup \mathcal{O}    
    \end{cases}$.
With $\mathcal{P}$ and $\mathcal{B}$ as input and $\mathcal{T}$ as output, a trained Numerical QA system can be regarded as a mapping $f$ such that
\begin{equation}
    f:\left( \mathcal{P},\mathcal{B}\right) \rightarrow \mathcal{T}
\end{equation}
Note that this expression not only describes the Numerical QA tasks, but also generalizes to other numeracy-related NLP tasks such as Tabular Entailment \citep{2019TabFactA} and Timeseries-based Fraudulent Detection \citep{padhi2021tabular}.

In this paper, we design and apply perturbations to the samples in the dataset to form perturbed prompt $\mathcal{P^\star}$, perturbed body $\mathcal{B^\star}$ and perturbed ground truth target $\mathcal{T^\star}$. We show that existing systems are fragile against numerical pertubation by showing that on a large portion of the dataset, the previous mapping fails to generate correct perturbed target, \ie:
\begin{equation}
\label{equ:perturbed}
    f:\left( \mathcal{P^\star},\mathcal{B^\star}\right) \not\rightarrow \mathcal{T^\star}
\end{equation}

%% file: body/4-Methodology.tex

\input{body/case_study}
Our approach aims at diagnosing the numerical weakness of existing Numerical Question Answering models. We list out and explain a series of \textbf{numerical capabilities} that are critical to solving Numerical Question Answering problems in \refsec{Numerical Capabilities}. We then design numerical \textbf{perturbations} targeting these capabilities in \refsec{Perturbations}. With the designed perturbations, we examine the weaknesses under two different \textbf{perturbations settings} in \refsec{Perturbing Settings}. 

These three sections are represented in \reffig{fig:overview}. as the ``\textbf{Capabilities}'' stripe, the ``\textbf{Perturbs}'' stripe, and the  ``\textbf{Attack Setting}'' and `` \textbf{Defense Setting}''.


\subsection{Numerical Capabilities}
\label{Numerical Capabilities}
We classify numerical capabilities into three major categories, concerning different aspects of numerical understanding, as below:

\textbf{Number Detection} is the capability of recognizing numbers of different surface forms. For instance, the English word \textit{"Forty-two"} and the Arabic number \textit{"42.0"} are regarded the same number in Numerical QA and should not affect the final arithmetic answer of a question. 

\textbf{Number Value Understanding} is the capability of understanding numbers of different value distributions. Systems are expected to not only apply arithmetic calculation on a specific set of numbers (\eg, integers of values smaller than 500 as included in the BERT tokenizer vocabulary). Robust Numerical QA systems are also expected to handle values such as float-point numbers and numbers larger than 500. 

\textbf{Operand Selection} is the capability of deciding which numbers to select as the operands in the arithmetic process. One important aspect of selecting related values is to exclude numbers that are 1) irrelevant to the Numerical QA problem scenario, or 2) relevant to the problem scenario but not essential to the question solving. Systems are expected to select as operands the important values from the unimportant values.

\textbf{Operation Reasoning} is the capability of inferring operations from the logic pattern described in the text. In an arithmetic process, the operation is independent from the operands, therefore different operations can be applied to the same set of selected related numbers in different questions. Systems are expected to decouple operation from operands and select the operation in an operand-agnostic way.

\subsection{Perturbations}
\label{Perturbations}
Perturbations are designed according to each numerical capabilities. In \reftab{tab:perturbed_examples}, an example problem is provided for each of the perturbations. The formal definition of the perturbations is provided in \refapp{sec:app_formal_definition}.

\textbf{Language Perturbation} targets the Number Detection capability and diagnoses how accurate can systems detect numbers in different surface forms. To perturb a number string $n_s$, we replace it with its English form of the number with Num2Words.\footnote{https://github.com/savoirfairelinux/num2words} This perturbation changes number surface forms but not their values. 

\textbf{Type Perturbation} targets the Number Detection capability and challenges systems to detect numbers in their float-point forms. To perturb a number string $n_s$, we concatenate it with the string ``.0''. Similar to Language Perturbation, only the number detection capability is diagnosed with this perturbation. Contrary to the Noise perturbation in the next paragraph, the Type perturbation does not propose additional calculation difficulty by changing number values.

\textbf{Noise Perturbation} targets the Number Value Understanding capability and challenges systems to not only understand arithmetic operations of not only integers but also float-point numbers. To perturb a number $n$, we randomly attach a one-digit fractional part with uniform distribution. This perturbation introduces new float-point numbers and breaks the original number value distribution in the dataset by adding an random variable. 

\textbf{Distribution Perturbation} targets the Number Value Understanding capability and challenges systems to conduct arithmetic with larger integers. To perturb a number $n$, we randomly offset the value with a normal distribution. Based on the observations in \citet{wallace-etal-2019-nlp}, we choose to perturb the majority of the numbers to larger than 500. This perturbation introduces large numbers and breaks original number value distribution in the dataset.

\textbf{Verbosity Perturbation} targets the Operand Selection capability and challenges systems to select the correct quantity in the problem by adding explicitly irrelevant numbers into the problem. To perturb a number string $n_s$, we concatenate it with an irrelevant number in parentheses, the irrelevant number is preceded by ``not''. This perturbation introduces numbers without breaking the distribution of relevant numbers in the dataset. 

\textbf{Extra Perturbation} targets the Operand Selection capability and challenges systems to exclude irrelevant numbers. To perturb a problem $(\mathcal{B}, \mathcal{P})$, An irrelvant sentence containing numbers randomly sampled from the corpus is added to the body $\mathcal{B}$. This perturbation breaks the number distribution by introducing extra instances of different numbers for the same problem. 

\textbf{Logic Perturbation} targets the Operation Reasoning capability and challenges systems to choose correct operations for the same set of numbers. In this paper, for two datasets described in \refsec{sec:exp_setup}, TATQA and ASDiv-a, the Operation perturbation demands additional attention. On TATQA it is based on template matching via SpaCy\footnote{https://spacy.io/} and automatic conversions, while on ASDiv-a it is based on manual annotation due to the diversity of patterns in the ASDiv-a dataset. This perturbation introduces extra problems of different operations. 
 
\textbf{Order Perturbation} targets the Operation Reasoning capability and challenges systems to choose correct operations for the same set of numbers. On ASDiv-a, the order of sentences in the problem body is manually altered in a manner that changes the order of number occurrence but not the problem logic. This perturbation does not break the operation distribution within the dataset. 

\subsection{Perturbing Settings}
\label{Perturbing Settings}
With the aforementioned perturbations, we construct perturbed datasets under different settings to investigate systems' numerical capabilities and the effectiveness of the perturbations from different perspectives. For a specific dataset with a training / validation / testing split, different splits are perturbed under different settings. In this paper we consider the following two settings of Attack and Defense, as compared in \reftab{tab:atk_def_diff}:

\textbf{Attack}. By applying the perturbations to the testing split of the dataset, we construct a challenge set to evaluate the corresponding numerical capability of existing systems. Systems are trained on the original datasets and evaluated on the perturbed challenge set. 

\textbf{Defense}. Under the defense setting, perturbations are applied to all of training, validation, and testing split of the dataset. By comparing systems' performance under the Defense with Attack settings, we investigate to what extent the performance drop can be alleviated by using the perturbations as a data augmentation approach. 

\input{body/atk_def_diff}

To perturb under Attack or Defense setting, suitable samples are first filtered according to a series of conditions. The perturbations are applied only to these filtered samples. The filtered samples in the dataset split(s) are replaced with their perturbed version to form the perturbed dataset. The filtering conditions and the formalized algorithm are provided in \refapp{sec:app_perturb_details}.


%% file: body/case_study.tex
\begin{table*}[!ht]
\centering
\resizebox{\textwidth}{!}{%
\begin{tabular}{@{}ccll@{}}
\toprule
\textbf{Capability} & \textbf{Perturbation} & \multicolumn{1}{c}{\textbf{Example Problem Pair}} & \multicolumn{1}{c}{\textbf{T5 Prediction}} \\ \midrule
\multirow{5}{*}{\begin{tabular}[c]{@{}c@{}}Number\\      Detection\end{tabular}} & Language & \begin{tabular}[c]{@{}l@{}}\original{Original:} A   mailman has to give out \original{192} pieces of junk mail. If he goes to \original{4} blocks, how \\  many pieces of junk mail should he give each block?\\      \perturbed{Perturbed:} A mailman has to give out \perturbed{one hundred and ninety-two} pieces of junk mail. If \\ he goes to \perturbed{four} blocks, how many pieces of junk mail should he give each block?\end{tabular} & \begin{tabular}[c]{@{}l@{}}\original{Original:}   192 / 4 \correcteq\\      \perturbed{Perturbed:} 92 / 4 \wrongeq\\ Expected: 192 / 4\end{tabular} \\ \cmidrule(l){2-4} 
 & Type & \begin{tabular}[c]{@{}l@{}}\original{Original:} There were \original{105} parents in the program and \original{698} pupils, too. How many people \\ were present in the program?\\      \perturbed{Perturbed:} There were \perturbed{105.0} parents in the program and \perturbed{698.0} pupils, too.   How many \\ people were present in the program?\end{tabular} & \begin{tabular}[c]{@{}l@{}}\original{Original:} 105 + 698 \correcteq\\      \perturbed{Perturbed:} 105 + 688 \wrongeq\\ Expected: 105 + 698\end{tabular} \\ \midrule
\multirow{5}{*}{\begin{tabular}[c]{@{}c@{}}Number Value\\      Understanding\end{tabular}} & Noise & \begin{tabular}[c]{@{}l@{}}\original{Original:} Tony   had \$\original{20}. He paid \$\original{8} for a ticket to a baseball game. At the game, he \\      bought a hot dog for \$\original{3}. What amount of money did Tony have then?\\      \perturbed{Perturbed:} Tony had \$\perturbed{20.2}. He paid \$\perturbed{8.5} for a ticket to a baseball game. At   the game, he \\bought a hot dog for \$\perturbed{3.5}. What amount of money did Tony have then?\end{tabular} & \begin{tabular}[c]{@{}l@{}}\original{Original:} 20 - 8 - 3 \correcteq\\      \perturbed{Perturbed:} 208.52 - 3.5 \wrongeq\\ Expected: 20.2 - 8.5 - 3.5 \end{tabular} \\ \cmidrule(l){2-4} 
 & Distribution & \begin{tabular}[c]{@{}l@{}}\original{Original:} Frank had \$\original{16}. After buying some new toys he had \$\original{8} left.How much did he\\ spend on toys?\\ \perturbed{Perturbed:} Frank had \$\perturbed{1281}. After buying some new toys he had \$\perturbed{478} left.How much did\\ he spend on toys?\end{tabular} & \begin{tabular}[c]{@{}l@{}}\original{Original:} 16 - 8 \correcteq\\      \perturbed{Perturbed:} 1215 - 878 \wrongeq\\ Expected: 1281 - 478\end{tabular} \\ \midrule
\multirow{5}{*}{\begin{tabular}[c]{@{}c@{}}Operand \\ Selection\end{tabular}} & Extra & \begin{tabular}[c]{@{}l@{}}\original{Original:} John   has twelve shirts. Later he bought four more shirts. How many shirts does \\ John have in total?\\      \perturbed{Perturbed:} John has twelve shirts. Later he bought four more shirts. \perturbed{Frank had \$16}. How \\ many shirts does John have in total?\end{tabular} & \begin{tabular}[c]{@{}l@{}}\original{Original:} 12 + 4 \correcteq\\      \perturbed{Perturbed:} 16 + 12 \wrongeq\\ Expected: 12 + 4\end{tabular} \\ \cmidrule(l){2-4} 
 & Verbosity & \begin{tabular}[c]{@{}l@{}}\original{Original:} The   roller coaster at the state fair costs \original{6} tickets per ride. If \original{8} friends were going \\ to ride the roller coaster, how many tickets would they need?\\      \perturbed{Perturbed:} The roller coaster at the state fair costs \perturbed{6 (not 30)} tickets   per ride. If \perturbed{8 (not 119)} \\ friends were going to ride the roller coaster, how   many tickets would they need?\end{tabular} & \begin{tabular}[c]{@{}l@{}}\original{Original:} 6 * 8 \correcteq\\      \perturbed{Perturbed:} 8 * 119 \wrongeq\\ Expected: 6 * 8\end{tabular} \\ \midrule
\multirow{5}{*}{\begin{tabular}[c]{@{}c@{}}Operation \\ Reasoning\end{tabular}} & Logic & \begin{tabular}[c]{@{}l@{}}\original{Original:} Jack received 8 emails in the morning and 2 emails in the afternoon. \original{How many} \\ \original{emails did Jack receive in the day?}\\      \perturbed{Perturbed:} Jack received 8 emails in the morning and 2 emails in the afternoon. \perturbed{How many} \\ \perturbed{more emails did Jack receive in the morning than in the afternoon?}\end{tabular} & \begin{tabular}[c]{@{}l@{}}\original{Original:} 8 + 2 \correcteq\\ \perturbed{Perturbed:} 8 + 2 \wrongeq\\ Expected: 8 - 2\end{tabular} \\ \cmidrule(l){2-4} 
 & Order & \begin{tabular}[c]{@{}l@{}}\original{Original:} \original{A DVD  book holds 126 DVDs. There are 81 DVDs already in the book.} How \\ many more DVDs can be put in the book?\\      \perturbed{Perturbed:} \perturbed{There are 81 DVDs already in the book. A DVD book holds 126 DVDs.} How \\many more DVDs can be put in the book?\end{tabular} & \begin{tabular}[c]{@{}l@{}}\original{Original:} 126 - 81 \correcteq\\      \perturbed{Perturbed:} 81 - 126 \wrongeq\\ Expected: 126 - 81\end{tabular} \\ \bottomrule
\end{tabular}%
}
\caption{Examples of DNC Perturbations and Corresponding Predictions by T5. For each perturbation an example \original{original} and \perturbed{perturbed} problem pair is shown. The rightmost column shows some error cases where T5 generates correct equation on the \original{original} problem but fails on the \perturbed{perturbed}. The ground truth equation of the \perturbed{perturbed} problem is also provided after ``Expected''.}
\label{tab:perturbed_examples}
\end{table*}

%% file: body/atk_def_diff.tex
\begin{table}[ht]
\centering
\resizebox{0.6\columnwidth}{!}{%
\begin{tabular}{@{}l|cc@{}}
\toprule
\textbf{Setting} & \textbf{Attack} & \textbf{Defense} \\ \midrule
\textbf{Train on} & train & train$^\star$ \\
\textbf{Validate   on} & val & val$^\star$ \\
\textbf{Test on} & test$^\star$ & test$^\star$ \\ \bottomrule
\end{tabular}%
}

\caption{The Comparison between Two Settings in DNC. Perturbations (denoted by ``$\star$'') are applied to different dataset splits (train / val / test) under each setting.}
\label{tab:atk_def_diff}
\end{table}

%% file: body/table_main_exp.tex
\begin{table*}[ht]
\centering
\resizebox{\textwidth}{!}{%
\begin{tabular}{@{}cc|cccccccc|cc|c@{}}
\toprule
\multicolumn{2}{c|}{} & \multicolumn{8}{c|}{ASDiv-a} & \multicolumn{2}{c|}{DROP-num} & TATQA-a \\ \cmidrule(l){3-13} 
\multicolumn{2}{c|}{\multirow{-2}{*}{Configuration}} & \multicolumn{2}{c}{T5} & \multicolumn{2}{c}{BART} & \multicolumn{2}{c}{GPT2} & \multicolumn{2}{c|}{Graph2Tree} & T5 & BART & TagOps \\ \midrule
Setting & Perturbation & Acc$_{eq}$ & Acc$_{ans}$ & Acc$_{eq}$ & Acc$_{ans}$ & Acc$_{eq}$ & Acc$_{ans}$ & Acc$_{eq}$ & Acc$_{ans}$ & Acc & Acc & Acc \\ \midrule
\multicolumn{1}{c|}{} & Language & \cellcolor[HTML]{BFD7EF}-18.85\% & \cellcolor[HTML]{BFD7EF}-18.85\% & \cellcolor[HTML]{AFCDEB}-23.77\% & \cellcolor[HTML]{A4C7E8}-27.05\% & \cellcolor[HTML]{D4E4F4}-12.30\% & \cellcolor[HTML]{D4E4F4}-12.30\% & \cellcolor[HTML]{E3EDF8}-7.65\% & \cellcolor[HTML]{E4EDF8}-7.38\% & \cellcolor[HTML]{D9E7F6}-10.62\% & \cellcolor[HTML]{CCDFF2}-14.73\% & \cellcolor[HTML]{C0D7EF}-18.62\% \\
\multicolumn{1}{c|}{} & Type & \cellcolor[HTML]{82B2DF}-37.70\% & \cellcolor[HTML]{D7E5F5}-11.48\% & \cellcolor[HTML]{92BCE3}-32.79\% & \cellcolor[HTML]{C9DDF1}-15.57\% & \cellcolor[HTML]{C4DAF0}-17.21\% & \cellcolor[HTML]{D9E7F6}-10.66\% & \cellcolor[HTML]{FCFCFF}0.27\% & \cellcolor[HTML]{FCFCFF}1.09\% & \cellcolor[HTML]{E3EDF8}-7.70\% & \cellcolor[HTML]{D8E6F5}-11.06\% & \cellcolor[HTML]{EAF1FA}-5.34\% \\
\multicolumn{1}{c|}{} & Noise & \cellcolor[HTML]{85B4E0}-36.89\% & \cellcolor[HTML]{85B4E0}-36.89\% & \cellcolor[HTML]{BFD7EF}-18.85\% & \cellcolor[HTML]{B7D2ED}-21.31\% & \cellcolor[HTML]{DCE8F6}-9.84\% & \cellcolor[HTML]{DEEAF7}-9.02\% & \cellcolor[HTML]{FCFCFF}0.27\% & \cellcolor[HTML]{FCFCFF}0.55\% & - & - & - \\
\multicolumn{1}{c|}{} & Distribution & \cellcolor[HTML]{C7DCF1}-16.39\% & \cellcolor[HTML]{CCDFF2}-14.75\% & \cellcolor[HTML]{9CC2E6}-29.51\% & \cellcolor[HTML]{C1D9EF}-18.03\% & \cellcolor[HTML]{D1E2F3}-13.11\% & \cellcolor[HTML]{D1E2F3}-13.11\% & \cellcolor[HTML]{E6EFF9}-6.56\% & \cellcolor[HTML]{E6EFF9}-6.56\% & - & - & - \\
\multicolumn{1}{c|}{} & Verbosity & \cellcolor[HTML]{75AADB}-41.80\% & \cellcolor[HTML]{6DA6D9}-44.26\% & \cellcolor[HTML]{AACAE9}-25.41\% & \cellcolor[HTML]{9CC2E6}-29.51\% & \cellcolor[HTML]{D9E7F6}-10.66\% & \cellcolor[HTML]{D7E5F5}-11.48\% & \cellcolor[HTML]{90BBE2}-33.33\% & \cellcolor[HTML]{8EBAE2}-33.88\% & \cellcolor[HTML]{DDE9F6}-9.58\% & \cellcolor[HTML]{D1E2F3}-13.31\% & \cellcolor[HTML]{F5F8FD}-1.90\% \\
\multicolumn{1}{c|}{} & Extra & \cellcolor[HTML]{AACAE9}-25.41\% & \cellcolor[HTML]{A2C5E7}-27.87\% & \cellcolor[HTML]{75AADB}-41.80\% & \cellcolor[HTML]{68A2D8}-45.90\% & \cellcolor[HTML]{9FC4E6}-28.69\% & \cellcolor[HTML]{9FC4E6}-28.69\% & \cellcolor[HTML]{5B9BD5}-53.83\% & \cellcolor[HTML]{5B9BD5}-54.64\% & \cellcolor[HTML]{D6E5F5}-11.79\% & \cellcolor[HTML]{D6E5F5}-11.67\% & \cellcolor[HTML]{F8F9FD}-1.21\% \\
\multicolumn{1}{c|}{} & Logic & \cellcolor[HTML]{9CC2E6}-29.51\% & \cellcolor[HTML]{A2C5E7}-27.87\% & \cellcolor[HTML]{85B4E0}-36.89\% & \cellcolor[HTML]{8AB7E1}-35.25\% & \cellcolor[HTML]{AACAE9}-25.41\% & \cellcolor[HTML]{AFCDEB}-23.77\% & \cellcolor[HTML]{A0C4E7}-28.42\% & \cellcolor[HTML]{B5D1EC}-21.86\% & - & - & \cellcolor[HTML]{CDE0F2}-14.29\% \\
\multicolumn{1}{c|}{\multirow{-8}{*}{Attack ($\Delta$)}} & Order & \cellcolor[HTML]{8DB9E2}-34.43\% & \cellcolor[HTML]{E9F0FA}-5.74\% & \cellcolor[HTML]{8FBAE2}-33.61\% & \cellcolor[HTML]{EEF4FB}-4.10\% & \cellcolor[HTML]{A2C5E7}-27.87\% & \cellcolor[HTML]{E4EDF8}-7.38\% & \cellcolor[HTML]{90BBE2}-33.33\% & \cellcolor[HTML]{E5EEF9}-7.10\% & - & - & \cellcolor[HTML]{FCFCFF}1.12\% \\ \midrule
\multicolumn{1}{c|}{} & Language & \cellcolor[HTML]{D4E4F4}-12.30\% & \cellcolor[HTML]{CFE0F3}-13.93\% & \cellcolor[HTML]{BCD5EE}-19.67\% & \cellcolor[HTML]{ACCCEA}-24.59\% & \cellcolor[HTML]{FCFCFF}2.46\% & \cellcolor[HTML]{FCFCFF}2.46\% & \cellcolor[HTML]{E3EDF8}-7.65\% & \cellcolor[HTML]{E4EDF8}-7.38\% & \cellcolor[HTML]{FCFCFF}0.07\% & \cellcolor[HTML]{F6F8FD}-1.84\% & \cellcolor[HTML]{E3EDF8}-7.59\% \\
\multicolumn{1}{c|}{} & Type & \cellcolor[HTML]{D7E5F5}-11.48\% & \cellcolor[HTML]{D4E4F4}-12.30\% & \cellcolor[HTML]{ECF2FA}-4.92\% & \cellcolor[HTML]{E6EFF9}-6.56\% & \cellcolor[HTML]{FCFCFF}3.28\% & \cellcolor[HTML]{FCFCFF}4.10\% & \cellcolor[HTML]{FCFCFF}1.64\% & \cellcolor[HTML]{FCFCFF}1.91\% & \cellcolor[HTML]{FCFCFF}0.46\% & \cellcolor[HTML]{F8FAFE}-0.95\% & \cellcolor[HTML]{FCFCFF}2.93\% \\
\multicolumn{1}{c|}{} & Noise & \cellcolor[HTML]{CCDFF2}-14.75\% & \cellcolor[HTML]{CCDFF2}-14.75\% & \cellcolor[HTML]{F1F5FC}-3.28\% & \cellcolor[HTML]{ECF2FA}-4.92\% & \cellcolor[HTML]{FCFCFF}3.28\% & \cellcolor[HTML]{FCFCFF}4.10\% & \cellcolor[HTML]{FCFCFF}0.55\% & \cellcolor[HTML]{FCFCFF}0.27\% & - & - & - \\
\multicolumn{1}{c|}{} & Distribution & \cellcolor[HTML]{BAD4ED}-20.49\% & \cellcolor[HTML]{BAD4ED}-20.49\% & \cellcolor[HTML]{E1ECF8}-8.20\% & \cellcolor[HTML]{DCE8F6}-9.84\% & \cellcolor[HTML]{E1ECF8}-8.20\% & \cellcolor[HTML]{DEEAF7}-9.02\% & \cellcolor[HTML]{E6EEF9}-6.83\% & \cellcolor[HTML]{E8F0F9}-6.01\% & - & - & - \\
\multicolumn{1}{c|}{} & Verbosity & \cellcolor[HTML]{C9DDF1}-15.57\% & \cellcolor[HTML]{C7DCF1}-16.39\% & \cellcolor[HTML]{E9F0FA}-5.74\% & \cellcolor[HTML]{E4EDF8}-7.38\% & \cellcolor[HTML]{F9FAFE}-0.82\% & \cellcolor[HTML]{FBFBFE}0.00\% & \cellcolor[HTML]{FBFBFE}-0.27\% & \cellcolor[HTML]{FCFCFF}1.09\% & \cellcolor[HTML]{EBF2FA}-5.13\% & \cellcolor[HTML]{F6F8FD}-1.84\% & \cellcolor[HTML]{FCFCFF}2.25\% \\
\multicolumn{1}{c|}{} & Extra & \cellcolor[HTML]{FCFCFF}0.00\% & \cellcolor[HTML]{FCFCFF}1.64\% & \cellcolor[HTML]{F4F7FC}-2.46\% & \cellcolor[HTML]{EEF4FB}-4.10\% & \cellcolor[HTML]{C4DAF0}-17.21\% & \cellcolor[HTML]{C1D9EF}-18.03\% & \cellcolor[HTML]{BAD4EE}-20.22\% & \cellcolor[HTML]{C2D9F0}-17.76\% & \cellcolor[HTML]{D7E6F5}-11.32\% & \cellcolor[HTML]{DAE7F6}-10.44\% & \cellcolor[HTML]{DEEAF7}-9.14\% \\
\multicolumn{1}{c|}{} & Logic & - & - & - & - & - & - & - & - & - & - & 13.64\% \\
\multicolumn{1}{c|}{\multirow{-8}{*}{Defense ($\Delta$)}} & Order & \cellcolor[HTML]{AACAE9}-25.41\% & \cellcolor[HTML]{EEF4FB}-4.10\% & \cellcolor[HTML]{A2C5E7}-27.87\% & \cellcolor[HTML]{E4EDF8}-7.38\% & \cellcolor[HTML]{F6F8FD}-1.64\% & \cellcolor[HTML]{FCFCFF}23.77\% & \cellcolor[HTML]{9DC3E6}-29.23\% & \cellcolor[HTML]{E2ECF8}-7.92\% & - & - & \cellcolor[HTML]{FCFCFF}19.47\% \\ \midrule
\multicolumn{1}{c|}{Original} & None & 68.03\% & 72.95\% & 67.21\% & 72.95\% & 44.26\% & 45.08\% & 66.94\% & 68.58\% & 49.42\% & 50.36\% & 42.41\% \\ \bottomrule
\end{tabular}%
}
\caption{The Results of DNC Framework. Five NLP systems are evaluated with three Numerical QA tasks under both Attack and Defense settings. The symbol ``$\Delta$'' stands for the absolute metric difference between the current setting and the original setting. The color scale represents the distance from the original setting, deeper means further from the original setting. For ASDiv-a, Acc$_{eq}$ and Acc$_{ans}$ refer to the prediction accuracy of ground truth equations and denotation accuracy of answers, respectively. For DROP-num and TATQA-a, Acc refers to the denotation accuracy of the answers. We provide the raw performance and relative change of the metrics \wrt the original setting in \refapp{sec:exp_abs_rel}. ``-'' denotes that automatic perturbation and automatic data augmentation as described by \refsec{Perturbing Settings} is not applicable here. We provide detailed explanation of the reason why they are not applicable in \refapp{sec:not_applicable}.}
\label{tab:main_exp}
\end{table*}

%% file: body/5-Experiments.tex

\subsection{Experiment Setup} 
\label{sec:exp_setup}
\input{body/table_dataset_stats}
\textbf{Datasets}. In this paper, we used ASDiv-a \citep{miao-etal-2020-diverse}, DROP \citep{dua-etal-2019-drop}, and TATQA \citep{zhu-etal-2021-tat} as our Numerical Question Answering datasets. For DROP and TATQA, we filtered out DROP-num and TATQA-a, the numerical subsets of them. The statistics of these datasets are shown in \reftab{tab:dataset_stats}.

\textbf{Systems}. We selected representative systems on each dataset and test their performance against perturbations. For the ASDiv-a dataset, we use Graph2Tree \cite{patel-etal-2021-nlp}. For the DROP dataset, we use BART-base and T5-base from Huggingface.\footnote{https://github.com/huggingface/transformers} For the TATQA dataset, we utilize TagOps with the RoBERTa backbone as described in the original paper. 

\textbf{Compute Environment}. All experiments are done on a Linux machine equipped with 4 NVIDIA Tesla V100 16GB GPUs. The average runtime of our experiments ranges from one to three hours. 

\textbf{Hyperparameters}. In our experiments, we adopt a general setting of hyperparameters of epoch number = $40$, learning rate = $1e-5$ and batch size = $32$. It is observed in our exploratory experiments that while the hyperparameters such as learning rate and batchsize do affect the absolute performance of the models, they have a modest effect on the general trend of the models’ strengths and weaknesses against the numerical perturbations. The details and analysis are provided in \refapp{sec:optim_param}.



\subsection{Experiment Results and Analysis}
\label{sec:exp_analysis}

The experiment results are provided in \reftab{tab:main_exp}. The metric we report is 1) the metric on original datasets (Original), and 2) the absolute change of the metric on perturbed datasets, denoted by ``$\Delta$''. We additionally provide the raw metric and relative drop in \reftab{tab:exp_orig} and \reftab{tab:exp_relative} in the Appendix. The calculation details of the observation can be found in \refapp{sec:calculation_details}.

\textbf{Attack}. As can be observed in \reftab{tab:main_exp} and \reftab{tab:exp_relative}, most systems were severely challenged under the Attack setting and experienced significant performance drop, ranging from 5\% to 50\% absolute drop and 5\% to 80\% relative drop in answer denotation accuracy. Between the two DNC goals, Semantic Parsing causes a more severe challenge, averaging \perfdrop{19.66}{31.79}, as compared to the \perfdrop{13.15}{19.66} by Numerical Parsing. 

Among the considered systems, Transformer-based Seq2Seq systems (T5, BART, GPT2) are more sensitive than the tasks-specific Graph2Tree system against the perturbations stemming from the Numerical Parsing goal. The former resulted in \perfdrop{17.42}{27.06}, while Graph2Tree only experienced \perfdrop{3.07}{4.48}. The masking of numbers used by Graph2Tree allows it to remain unaffected against a portion of the perturbations targeting the Numerical Parsing goal.

\textbf{Defense}. As a counteracting approach, the defense mechanism helps alleviate systems' lack of corresponding numerical capabilities by applying automatic perturbations to the training and validation set. Via Defense, the lack according to the Semantic Parsing gets more recovery of (\perfimprove{17.96}{26.95} \vs \perfimprove{6.52}{11.42}). 

Among the considered systems, Transformer-based Seq2Seq systems benefits more from Defense than the Graph2Tree system (\perfimprove{12.53}{20.52} \vs \perfimprove{11.58}{16.88}).

Despite the recovery from Defense, the challenge is still not solved. As the majority of the defense performance is still more than 10\% below the original performance. This observation indicates that the lack of Numerical Capabilities is still an open question.

\textbf{Summary}. Our DNC framework provides insights on two major aspects of the diagnosis to Numerical QA systems:

1) It is demonstrated that severe numerical weaknesses exist in current Numerical QA systems (``Attack''), and they can not be trivially eliminated via, although benefiting from, an automatic data augmentation process (``Defense'').

2) The systems' weaknesses are explicitly profiled in a quantitative and interpretable manner through the models' susceptibility difference to a diversity of perturbations.

%% file: body/table_dataset_stats.tex
\begin{table}[ht]
\centering
\resizebox{\columnwidth}{!}{%
\begin{tabular}{@{}cccc@{}}
\toprule
\textbf{Dataset} & \textbf{\# Training} & \textbf{\# Validation} & \textbf{\# Testing} \\ \midrule
ASDiv-a & 974 & 122 & 122 \\
DROP-num & 42258 & 5282 & 5283 \\
TATQA-a & 1971 & 245 & 247 \\ \bottomrule
\end{tabular}%
}
\caption{The Statistics of the Datasets Used.}
\label{tab:dataset_stats}
\vspace{-5mm}
\end{table}

%% file: body/6-Potential_Directions.tex
In this section, phenomena observed on different systems and datasets were summarized to provide comparison for existing methods. Also, recent related efforts corresponding to these phenomena were discussed to point open directions in the domain of Numerical QA.
\subsection{Target: Logical Form Generation \vs Answer Predicting}
\label{sec:eq_vs_ans}
One attribute specific to Numerical QA is the reasoning processes leading to the numerical answers, which is usually described by logical forms. On datasets where the ground truth logical forms are provided as an additional supervision (\eg, ASDiv-a and TATQA), the systems have two options for the target: 1) \textbf{Logical Form Generation}, where systems generate the logical form which is later input to external symbolic executing systems such as Python scripts or SQL engines, and 2) \textbf{Answer Predicting}, where systems directly predict the output answer in an end-to-end manner. On datasets where ground truth logical forms are not provided (\eg, DROP), the latter is the most frequently adopted approach. Logical Form Generation and Answer Predicting differ in the actual object to conduct the executing step of the logical form insinuated by the question (external symbolic systems \vs neural systems). With Answer Predicting, systems are expected to possess the capability of executing the logical forms internally. 

We investigate to what extent do existing systems possess this execution capability, by comparing the impact of the problem target $\mathcal{T}$ in Numerical QA on ASDiv-a. The systems are trained to predict two different targets: 1) the logical form (\ie, the MWP equation), and 2) the logical form and the execution result. Since most MWP-specific systems are incapable of predicting answers directly, we choose the Transformer-based systems GPT2, BART and T5. Results in \reftab{tab:emp_compare_models} indicate that: 1) on existing systems, Logical Form Generation is beneficial for higher accuracy, and 2) even though models managed to compose equations with high accuracy, they struggle to faithfully execute an equation to get the correct answer.

\input{body/table_emp_compare_models}


Recent work also pays increasing attention to the execution capability. Systems such as TAPEX \citep{liu2022tapex} and POET \citep{Pi2022ReasoningLP} have been leveraging data synthesizing and intermediate pretraining to learn neural program executors and achieved state-of-the-art results over systems leveraging Logical Form Generation. This recent development shows the potential of neural systems with enhanced execution capability on the Numerical QA task.

\subsection{Numbers: Tokenization \vs Replacement}
\label{sec:tok_vs_mask}

We also investigate the impact of different ways of manipulating numbers. There are two mainstream existing methods to process and represent numbers, herein referred to as the \textbf{Tokenization} and \textbf{Replacement} methods.

Tokenization methods such as WordPiece \citep{Wu2016GooglesNM} and BPE \citep{sennrich-etal-2016-neural} adopted by existing Numerical QA systems divides numbers into potentially multiple sub-word level tokens. \Eg, The number \texttt{768} will be divided into tokens \texttt{7} and \texttt{68} by T5's tokenizer. This approach stems from the fundamental fact that existing systems' vocabularies are finite while the occurrences of numbers in a Numerical QA dataset can be too diverse to include in a finite vocabulary. Tokenization causes extra representation cost and erases the digit integrity by potentially introducing multiple tokens for a single number. 

Replacement substitutes numbers with special tokens in the input (\texttt{[NUM1]}, \texttt{[NUM2]}, \etc), which are later re-substituted with the original number in the output logical forms. This approach avoids multiple tokens by providing exactly one representation for each number, but has its own limitations handling number diversity since the recognition of numbers are usually performed with rule-based matching, which is often non-exhaustive.

In this paper, T5, BART, GPT2 and TagOps adopts Tokenization, while Graph2Tree adopts Replacement. We implement two variations of GPT2: GPT2$_{token}$ and GPT2$_{replace}$ to compare their robustness against different perturbations on the ASDiv-a dataset.Results in \reftab{tab:emp_compare_handling} indicate that Replacement has an advantage when no perturbation is present or when the perturbation only involves changes in number value. However, when the perturbation changes number values, the Replacement-based system is more severely challenged.

\input{body/table_emp_compare_handling}

We hypothesize that the Replacement method removes all numerical information such as the format and value of numbers in the problem and lost numeracy capabilities, therefore the system receives only textual signals such as number order or word frequency, which further encouraged systems to learn from spurious correlations as stated in \citet{patel-etal-2021-nlp}. This hypothesis is consistent with the observations of a recent study \citep{thawani-etal-2021-numeracy} that investigates of the mutual-enhancement between numeracy and literacy.

The respective limitations of Tokenization and Replacement are calling for more numeracy-preserving number representation methods. Some studies have suggested changing number surface forms \cite{kim-etal-2021-seen} or using dataset-agnostic representation \citep{sundararaman-etal-2020-methods}, however they either create extra token loads or could not generalize well on large-scale real-world dataset. The numeracy-preserving number representation is another bottleneck for Numerical QA.

%% file: body/table_emp_compare_models.tex
\begin{table}[t]
\centering
\resizebox{0.5\columnwidth}{!}{%
\begin{tabular}{@{}lcc@{}}
\toprule
\multicolumn{1}{c}{\textbf{Model}} & \textbf{Acc$_{ans}$} & \textbf{Acc$_{eq}$} \\ \midrule
GPT2$_{ans}$ & 6.56\% & - \\
GPT2$_{eq}$ & 45.08\% & 44.26\% \\ \midrule
BART$_{ans}$ & 9.02\% & - \\
BART$_{eq}$ & 72.95\% & 67.21\% \\ \midrule
T5$_{ans}$ & 2.46\% & - \\
T5$_{eq}$ & 72.95\% & 68.03\% \\ \bottomrule
\end{tabular}%
}
\caption{Comparing Models with Different Prediction Targets on ASDiv-a. For a model $\mathcal{M}$, $\mathcal{M}_{eq}$ / $\mathcal{M}_{ans}$ predicts equation / equation and answer, respectively. Acc$_{eq}$ and Acc$_{ans}$ stand for the denotation accuracy of the generated equation and the accuracy of the directly predicted answer, respectively.}
\label{tab:emp_compare_models}
\vspace{-5mm}
\end{table}

%% file: body/table_emp_compare_handling.tex
\begin{table}[t]
\centering
\resizebox{\columnwidth}{!}{%
\begin{tabular}{@{}cccccc@{}}
\toprule
\textbf{Model} & \textbf{Perturbation} & \textbf{Acc$_{eq}$} & \textbf{Acc$_{ans}$} & \textbf{$\Delta$ Acc$_{eq}$} & \textbf{$\Delta$ Acc$_{ans}$} \\ \midrule
 & Language & 31.97\% & 32.79\% & \cellcolor[HTML]{D6E6F4}-12.30\% & \cellcolor[HTML]{D6E6F4}-12.30\% \\
 & Type & 27.05\% & 34.43\% & \cellcolor[HTML]{C6DCF0}-17.21\% & \cellcolor[HTML]{DCE9F6}-10.66\% \\
 & Noise & 34.43\% & 36.07\% & \cellcolor[HTML]{DEEBF6}-9.84\% & \cellcolor[HTML]{E1ECF7}-9.02\% \\
 & Distribution & 31.15\% & 31.97\% & \cellcolor[HTML]{D3E4F3}-13.11\% & \cellcolor[HTML]{D3E4F3}-13.11\% \\
 & Verbosity & 33.61\% & 33.61\% & \cellcolor[HTML]{DCE9F6}-10.66\% & \cellcolor[HTML]{D9E8F5}-11.48\% \\
 & Extra & 15.57\% & 16.39\% & \cellcolor[HTML]{A0C5E6}-28.69\% & \cellcolor[HTML]{A0C5E6}-28.69\% \\
 & Logic & 18.85\% & 21.31\% & \cellcolor[HTML]{ABCCE9}-25.41\% & \cellcolor[HTML]{B1CFEB}-23.77\% \\
 & Order & 16.39\% & 37.70\% & \cellcolor[HTML]{A3C7E7}-27.87\% & \cellcolor[HTML]{E6F0F8}-7.38\% \\
\multirow{-9}{*}{GPT2$_{token}$} & Original & 44.26\% & 45.08\% & \cellcolor[HTML]{FFFFFF}0.00\% & \cellcolor[HTML]{FFFFFF}0.00\% \\ \midrule
 & Language & 46.72\% & 47.54\% & \cellcolor[HTML]{DEEBF6}-9.84\% & \cellcolor[HTML]{DEEBF6}-9.84\% \\
 & Type & 56.56\% & 57.38\% & \cellcolor[HTML]{FFFFFF}0.00\% & \cellcolor[HTML]{FFFFFF}0.00\% \\
 & Noise & 56.56\% & 57.38\% & \cellcolor[HTML]{FFFFFF}0.00\% & \cellcolor[HTML]{FFFFFF}0.00\% \\
 & Distribution & 56.56\% & 57.38\% & \cellcolor[HTML]{FFFFFF}0.00\% & \cellcolor[HTML]{FFFFFF}0.00\% \\
 & Verbosity & 22.13\% & 22.13\% & \cellcolor[HTML]{8EBAE2}-34.43\% & \cellcolor[HTML]{8BB8E1}-35.25\% \\
 & Extra & 10.66\% & 11.48\% & \cellcolor[HTML]{68A3D8}-45.90\% & \cellcolor[HTML]{68A3D8}-45.90\% \\
 & Logic & 32.79\% & 40.16\% & \cellcolor[HTML]{B1CFEB}-23.77\% & \cellcolor[HTML]{C6DCF0}-17.21\% \\
 & Order & 18.03\% & 36.07\% & \cellcolor[HTML]{80B1DE}-38.52\% & \cellcolor[HTML]{B9D4ED}-21.31\% \\
\multirow{-9}{*}{GPT2$_{replace}$} & Original & 56.56\% & 57.38\% & \cellcolor[HTML]{FFFFFF}0.00\% & \cellcolor[HTML]{FFFFFF}0.00\% \\ \bottomrule
\end{tabular}%
}
\caption{The Results of Tokenization and Replacement on GPT2. GPT2$_{token}$ adopts the Tokenization method and GPT2$_{replace}$ adopts the Replacement method.}
\label{tab:emp_compare_handling}
\end{table}

%% file: body/7-Conclusion.tex
In this paper we aim at diagnosing numerical capabilities in existing NLP systems. We list out a series of numerical capabilities and design corresponding dataset perturbations. Empirical results show that existing systems still lack numerical capabilities to a large extent, and this lack cannot be eliminated in a trivial manner. Analysis into the empirical results, discussion of the the existing practices, and insights for future directions of Numerical QA dataset collection and system design are also provided.

%% file: body/Appendix.tex
\section{Formal Definition of Perturbations}
\label{sec:app_formal_definition}
We provide the formalized definition of the perturbations as follows. In all definitions, ``$\star$'' denotes perturbed version.

\textbf{Noise Perturbation}.
To apply noise perturbation to an number $n$, an variable $X$ is uniformly sampled on the interval (1, 10). Then a fractional part corresponding to $X$ is added the concerned number $n$, \ie,
\begin{align*}
    X &\sim \mathcal{U}(1,10) \\
    n^\star &= n + 0.1 \times \lfloor X \rfloor
\end{align*}

\textbf{Distribution Perturbation}.
The Distribution Perturbation changes the number distribution in the dataset by adding an normally distributed random variable $X$ to the concerned number $n$. \Ie,
\begin{align*}
    X &\sim \mathcal{N}(\mu_d, \delta_d^2) \\
    n^\star &= n + \lfloor X \rfloor
\end{align*}
In this paper we adopt $\mu_d = 1000$ and $\delta_d = 300$.

\textbf{Language Perturbation}.
The concerned number string $n_s$ is replaced by the English word describing the same quantity, \ie,
\begin{equation*}
    n_s^\star = Num2Words(n_s)
\end{equation*}

\textbf{Type Perturbation}.
To apply the Type Perturbation, the concerned number is expected to be an integral number. The number string $n_s$ is concatenated with an extra ``.0'' string to change the type of the concerned number from integer to float-point, \ie,
\begin{equation*}
    n_s^\star = Concat(n_s, Stringfy(.0))
\end{equation*}

\textbf{Verbosity Perturbation}.
The Verbosity Perturbation aims to introduce irrelevant numbers without changing the semantics of the problem. To perturb a number string $n_s$, we concatenate it with an irrelevant number in parentheses, the irrelevant number is preceded by "not", \ie,
\begin{align*}
    X &\sim \mathcal{N}(\mu_v, \delta_v^2) \\
    n_s^\star &= Concat(ns, (not, Stringfy(X))) \\
\end{align*}
In this paper we adopt $\mu_v = 100$ and $\delta_v=30$.

\textbf{Extra Perturbation}
To apply the Extra Perturbation to a problem $(\mathcal{B}, \mathcal{P})$, an irrelevant sentence containing numbers from the corpus is added to the body $\mathcal{B}$, \ie,
\begin{align*}
    \mathcal{P}_\Delta &= SampleOtherQs() \\
    \mathcal{P}^\star &= \mathcal{P} \oplus \mathcal{P}_\Delta  \\
    \mathcal{B}^\star &= \mathcal{B}
\end{align*}

\textbf{Logic Perturbation}
To apply the Logic Perturbation to a problem $(\mathcal{B}, \mathcal{P})$, the prompt is altered to convert the problem logic used in the problem, \ie,
\begin{align*}
    \mathcal{P}^\star &= ConvertLogic(\mathcal{P}) \\
    \mathcal{B}^\star &= \mathcal{B}
\end{align*}

\textbf{Order Perturbation}
For the Order Perturbation, the sentence order in the problem body is manually altered in a manner that changes the order of number occurrence but not the problem logic, \ie,
\begin{align*}
    \mathcal{P}^\star &= ChangeOrder(\mathcal{P}) \\
    \mathcal{B}^\star &= \mathcal{B}
\end{align*}


\section{Details of the Perturbing Process} 
\label{sec:app_perturb_details}

\subsection{The Filtering Conditions}
The filtering conditions for Perturbing Algorithms is different across perturbations. The perturbations can be divided into two major categories: 1) perturbations that do not change the solving equation or final results (Language, Type, Verbosity, Extra, Order), and 2) perturbations that changes the solving equation or final results (Noise, Distri, Operation). 

For perturbations in category 1), there is no limitation on the perturbing process, thus all questions naturally pass the the filtering condition. 

For perturbations in category 2), the filtering conditions follow the principles of Unambiguity, Suitability and Visibility.

\textbf{Unambiguity}. The filtered question should have an unambiguous mapping between the number to be perturbed and the their location in the context. One example that violates this principle is when there are duplicated numbers in the problem body, then it cannot be determined which occurrence of the number affects the final result.

\textbf{Suitability}. The number to be perturbed should be suitable for the perturbation to be conducted. \Eg A float-point number should not be used as the target of the Noise perturbation which adds fractional part to integral numbers. In DNC, the Noise and Type perturbations requires the concerned number to be integral, and the Operation perturbation requires the question to match a manually created template.

\textbf{Visibility}. The concerned number should be occur in the the problem since the perturbations can only be applied to known input numbers.

\subsection{The Formalized Perturbing Process}
\input{body/att_algo}

\section{Hyperparameters}
\label{sec:optim_param}
In our exploratory experiment, it it observed that while the hyperparameters such as learning rate and batchsize do affect the absolute performance of the models, they have a modest effect on the general trend of the models’ strengths and weaknesses against the numerical perturbations. We hypothesize that this is due to the numerical capabilities of a model being contributed mostly by the model architecture instead of hyperparameters. 

For example, when the hyperparameters are varying from the default setting (1e-5 for learning rate, 32 for batch size), the following results are observed: 

On Graph2Tree, the results of changing the learning rate and batch size are shown in \reftab{tab:g2t_hp}, the trend of results with the varied hyperparameters align with the default result as shown in \reftab{tab:main_exp}. 

\begin{table}[h]
\resizebox{\columnwidth}{!}{%
\begin{tabular}{@{}cccc@{}}
\toprule
Perturbation & \begin{tabular}[c]{@{}c@{}}$Acc_{eq}$\\      lr=1e-2\end{tabular} & \begin{tabular}[c]{@{}c@{}}$Acc_{eq}$\\      batchsize=128\end{tabular} & Default                          \\ \midrule
Language     & \cellcolor[HTML]{E0EBF7}-4.10\%                                   & \cellcolor[HTML]{D9E7F5}-9.84\%                                         & \cellcolor[HTML]{E4EDF8}-7.65\%  \\
Type         & \cellcolor[HTML]{F9FAFE}3.28\%                                    & \cellcolor[HTML]{FCFCFF}3.28\%                                          & \cellcolor[HTML]{FCFCFF}0.27\%   \\
Noise        & \cellcolor[HTML]{FCFCFF}4.10\%                                    & \cellcolor[HTML]{F7F9FD}1.64\%                                          & \cellcolor[HTML]{FCFCFF}0.27\%   \\
Distribution & \cellcolor[HTML]{DDE9F7}-4.92\%                                   & \cellcolor[HTML]{D7E5F5}-10.66\%                                        & \cellcolor[HTML]{E7EFF9}-6.56\%  \\
Verbosity    & \cellcolor[HTML]{7EB0DE}-33.61\%                                  & \cellcolor[HTML]{8DB9E2}-38.52\%                                        & \cellcolor[HTML]{98BFE4}-33.33\% \\
Extra        & \cellcolor[HTML]{5B9BD5}-44.26\%                                  & \cellcolor[HTML]{5B9BD5}-57.38\%                                        & \cellcolor[HTML]{5B9BD5}-53.83\% \\
Logic        & \cellcolor[HTML]{B2CFEB}-18.03\%                                  & \cellcolor[HTML]{A9CAE9}-27.87\%                                        & \cellcolor[HTML]{A6C8E8}-28.42\% \\
Order        & \cellcolor[HTML]{7EB0DE}-33.61\%                                  & \cellcolor[HTML]{D0E1F3}-13.11\%                                        & \cellcolor[HTML]{98BFE4}-33.33\% \\
Original     & 62.30\%                                                           & 71.31\%                                                                 & 66.94\%                          \\ \bottomrule
\end{tabular}%
}
\caption{The results of the attack $Acc_{eq}$ for Graph2Tree on ASDiv-a}
\label{tab:g2t_hp}
\end{table}

Similar behavior can also be observed on large transformer-based model such as T5, as shown in \reftab{tab:t5_hp}: 
\begin{table}[h]
\resizebox{\columnwidth}{!}{%
\begin{tabular}{@{}cccc@{}}
\toprule
Perturbation & \begin{tabular}[c]{@{}c@{}}$Acc_{eq}$\\ lr=1e-4\end{tabular} & \begin{tabular}[c]{@{}c@{}}$Acc_{eq}$\\ batchsize=16\end{tabular} & Default                          \\ \midrule
Language     & \cellcolor[HTML]{FCFCFF}-17.21\%                             & \cellcolor[HTML]{E2ECF8}-17.21\%                                  & \cellcolor[HTML]{ECF2FA}-18.85\% \\
Type         & \cellcolor[HTML]{E4EEF8}-20.49\%                             & \cellcolor[HTML]{7FB0DE}-32.79\%                                  & \cellcolor[HTML]{74AADB}-37.70\% \\
Noise        & \cellcolor[HTML]{CDE0F2}-23.77\%                             & \cellcolor[HTML]{7FB0DE}-32.79\%                                  & \cellcolor[HTML]{7AADDD}-36.89\% \\
Distribution & \cellcolor[HTML]{F6F8FD}-18.03\%                             & \cellcolor[HTML]{FCFCFF}-13.11\%                                  & \cellcolor[HTML]{FCFCFF}-16.39\% \\
Verbosity    & \cellcolor[HTML]{609ED6}-39.34\%                             & \cellcolor[HTML]{5B9BD5}-38.52\%                                  & \cellcolor[HTML]{5B9BD5}-41.80\% \\
Extra        & \cellcolor[HTML]{5B9BD5}-40.16\%                             & \cellcolor[HTML]{C8DCF1}-21.31\%                                  & \cellcolor[HTML]{C2D9F0}-25.41\% \\
Logic        & \cellcolor[HTML]{B6D2EC}-27.05\%                             & \cellcolor[HTML]{99C0E5}-28.69\%                                  & \cellcolor[HTML]{A8C9E9}-29.51\% \\
Order        & \cellcolor[HTML]{9AC1E5}-31.15\%                             & \cellcolor[HTML]{7AADDD}-33.61\%                                  & \cellcolor[HTML]{89B7E1}-34.43\% \\
Original     & 61.48\%                                                      & 63.11\%                                                           & 68.03\%                          \\ \bottomrule
\end{tabular}%
}
\caption{The results of the attack $Acc_{eq}$ for T5 on ASDiv-a}
\label{tab:t5_hp}
\end{table}

Considering this observation, and the fact that the number of our experiment is large due to the combination of different models, datasets, DNC settings, and DNC perturbations, we chose one general setting to reduce search space. We chose the setting as close as possible to the reported setting in the original papers of Graph2Tree and T5. We verified that this setting provides sufficiently good performance to demonstrate the performance gap corresponding to the perturbations, since our experiment focused more on the performance of a same model checkpoint against the datasets before and after the perturbations.

\section{DNC Results}
\subsection{Raw Performance And Relative Performance Drop}
\label{sec:exp_abs_rel}
We provide the original result in \reftab{tab:exp_orig} and the relative performance drop in \reftab{tab:exp_relative}. 

\subsection{Observation Calculation Details}
We denote the experiment results table in \reftab{tab:main_exp_with_coord}. The values, observation explanation, and the formula used are provided in \reftab{tab:exp_formula}.
\label{sec:calculation_details}

\section{DNC Experiments that Are Not Applicable}
\label{sec:not_applicable}
The following types of experiments are not applicable in current DNC framework:
\subsection{The Defense of the Logic perturbation on ASDiv-a}
The Logic perturbation requires the problem to be perturbed in a way that the logic is changed while the semantics of the problem is still cohesive. This requirement proposes challenge on the scalability of the perturbation. For the Attack setting, we utilized manually annotated labels. However, under the Defense setting the perturbations are expected to automatically augment the dataset. Thus, the Defense setting results of Logic perturbation on ASDiv-a is not applicable.

\subsection{Noise and Type perturbations on DROP-num and TATQA-a}
DROP-num and TATQA-a do not provide supervision of the operand origins, therefore a mapping from the operands in equation to the context quantities cannot be built, which results in the Noise and Distribution perturbation not applicable on the DROP-num and TATQA-a datasets.

\subsection{Logic Perturbation on DROP-num}
DROP-num does not provide ground truth reasoning steps or logical forms, thus Logic perturbations that has dependency on the provided supervision is not applicable on DROP-num.

\subsection{Order Perturbation on DROP-num}
DROP-num is a reasonding dataset based on real-world paragraphs that usually have logical or temporal order information. Order perturbation breaks the semantic of the paragraph and will also confuse humans. Thus Order perturbation is not valid on DROP-num and the results are not applicable.

\clearpage

\input{body/table_exp_orig}
\input{body/table_exp_relative}
\input{body/table_main_exp_with_coord}

%% file: body/att_algo.tex
\SetKwComment{Comment}{/* }{ */}
\RestyleAlgo{ruled}
\SetKwComment{Comment}{/* }{ */}
\begin{algorithm}[ht]
\caption{The Perturbing Process}
\label{attack_algorithm}
\KwData{$D = \{D_{train}, D_{val}, D_{test}\}$ 
        $AllPert = \{Noise, ...\}$ as in \refsec{Perturbations}
        $AllSet \in \{Attack, ...\}$ as in \refsec{Perturbing Settings}
        $Filter = \{Integral, ...\}$ as in \refsec{sec:app_perturb_details}} 
\KwResult{$D^\star = \{D^\star_{train}, D^\star_{val}, D^\star_{test}\}$}
\Comment{Decide perturbations to use}
$Perturbs \gets SelectBy(AllPert, D)$\;
\Comment{Create perturbed dataset for each perturbation and setting}
\For{$setting \in AllSet$}{
\Comment{Decide split to perturb}
$D_{\Delta}, D_{remain} \gets SelectBy(D, setting)$\;
    \For{$perturb \in Perturbs$}{
        $D^\star_{\Delta} \gets \{\}$\;
        \For{$d \in D_{\Delta}$}{
            \If{Filter(d)}{
                $D^\star_{\Delta} \gets D^\star_{\Delta} + perturb(d)$\;
            }
            \Else{
                $D^\star_{\Delta} \gets D^\star_{\Delta} + d$\;
            }
        }
        $D^\star_{perturb} = \{D^\star_{\Delta}, D_{remain}\}$
    }
}
$D^\star = \{D^\star_{perturb}\}|~_{perturb ~\in~ Perturbs}$
\end{algorithm}

%% file: body/table_exp_orig.tex
\begin{table*}[ht]
\centering
\resizebox{\textwidth}{!}{%
\begin{tabular}{@{}cc|cccccccc|cc|c@{}}
\toprule
\multicolumn{2}{c|}{\multirow{2}{*}{Configuration}} & \multicolumn{8}{c|}{ASDiv-a} & \multicolumn{2}{c|}{DROP-num} & TATQA-a \\ \cmidrule(l){3-13} 
\multicolumn{2}{c|}{} & \multicolumn{2}{c}{T5} & \multicolumn{2}{c}{BART} & \multicolumn{2}{c}{GPT2} & \multicolumn{2}{c|}{Graph2Tree} & T5 & BART & TagOps \\ \midrule
\multicolumn{1}{c|}{Setting} & Perturbation & Acc$_{eq}$ & Acc$_{ans}$ & Acc$_{eq}$ & Acc$_{ans}$ & Acc$_{eq}$ & Acc$_{ans}$ & Acc$_{eq}$ & Acc$_{ans}$ & Acc & Acc & Acc \\ \midrule
\multicolumn{1}{c|}{\multirow{8}{*}{Attack}} & Language & 49.18\% & 54.10\% & 43.44\% & 45.90\% & 31.97\% & 32.79\% & 59.29\% & 61.20\% & 38.80\% & 35.62\% & 23.79\% \\
\multicolumn{1}{c|}{} & Type & 30.33\% & 61.48\% & 34.43\% & 57.38\% & 27.05\% & 34.43\% & 67.21\% & 69.67\% & 41.72\% & 39.30\% & 37.07\% \\
\multicolumn{1}{c|}{} & Noise & 31.15\% & 36.07\% & 48.36\% & 51.64\% & 34.43\% & 36.07\% & 67.21\% & 69.13\% & - & - & - \\
\multicolumn{1}{c|}{} & Distribution & 51.64\% & 58.20\% & 37.70\% & 54.92\% & 31.15\% & 31.97\% & 60.38\% & 62.02\% & - & - & - \\
\multicolumn{1}{c|}{} & Verbosity & 26.23\% & 28.69\% & 41.80\% & 43.44\% & 33.61\% & 33.61\% & 33.61\% & 34.70\% & 39.84\% & 37.04\% & 40.52\% \\
\multicolumn{1}{c|}{} & Extra & 42.62\% & 45.08\% & 25.41\% & 27.05\% & 15.57\% & 16.39\% & 13.11\% & 13.93\% & 37.63\% & 38.69\% & 41.21\% \\
\multicolumn{1}{c|}{} & Logic & 38.52\% & 45.08\% & 30.33\% & 37.70\% & 18.85\% & 21.31\% & 38.52\% & 46.72\% & - & - & 28.12\% \\
\multicolumn{1}{c|}{} & Order & 33.61\% & 67.21\% & 33.61\% & 68.85\% & 16.39\% & 37.70\% & 33.61\% & 61.48\% & - & - & 43.53\% \\ \midrule
\multicolumn{1}{c|}{\multirow{8}{*}{Defense}} & Language & 55.74\% & 59.02\% & 47.54\% & 48.36\% & 46.72\% & 47.54\% & 59.29\% & 61.20\% & 49.49\% & 48.52\% & 34.83\% \\
\multicolumn{1}{c|}{} & Type & 56.56\% & 60.66\% & 62.30\% & 66.39\% & 47.54\% & 49.18\% & 68.58\% & 70.49\% & 49.88\% & 49.41\% & 45.34\% \\
\multicolumn{1}{c|}{} & Noise & 53.28\% & 58.20\% & 63.93\% & 68.03\% & 47.54\% & 49.18\% & 67.49\% & 68.85\% & - & - & - \\
\multicolumn{1}{c|}{} & Distribution & 47.54\% & 52.46\% & 59.02\% & 63.11\% & 36.07\% & 36.07\% & 60.11\% & 62.57\% & - & - & - \\
\multicolumn{1}{c|}{} & Verbosity & 52.46\% & 56.56\% & 61.48\% & 65.57\% & 43.44\% & 45.08\% & 66.67\% & 69.67\% & 44.29\% & 48.52\% & 44.67\% \\
\multicolumn{1}{c|}{} & Extra & 68.03\% & 74.59\% & 64.75\% & 68.85\% & 27.05\% & 27.05\% & 46.72\% & 50.82\% & 38.10\% & 39.92\% & 33.28\% \\
\multicolumn{1}{c|}{} & Logic & - & - & - & - & - & - & - & - & - & - & 56.05\% \\
\multicolumn{1}{c|}{} & Order & 42.62\% & 68.85\% & 39.34\% & 65.57\% & 42.62\% & 68.85\% & 37.70\% & 60.66\% & - & - & 61.89\% \\ \midrule
\multicolumn{1}{c|}{Original} & None & 68.03\% & 72.95\% & 67.21\% & 72.95\% & 44.26\% & 45.08\% & 66.94\% & 68.58\% & 49.42\% & 50.36\% & 42.41\% \\ \bottomrule
\end{tabular}%
}
\caption{The raw results of the DNC Framework. Notations here follow the ones in the main experiment results}
\label{tab:exp_orig}
\end{table*}

%% file: body/table_exp_relative.tex
\begin{table*}[ht]
\centering
\resizebox{\textwidth}{!}{%
\begin{tabular}{@{}cc|cccccccc|cc|c@{}}
\toprule
\multicolumn{2}{c|}{} & \multicolumn{8}{c|}{ASDiv-a} & \multicolumn{2}{c|}{DROP-num} & TATQA-a \\ \cmidrule(l){3-13} 
\multicolumn{2}{c|}{\multirow{-2}{*}{Configuration}} & \multicolumn{2}{c}{T5} & \multicolumn{2}{c}{BART} & \multicolumn{2}{c}{GPT2} & \multicolumn{2}{c|}{Graph2Tree} & T5 & BART & TagOps \\ \midrule
Setting & Perturbation & Acc$_{eq}$ & Acc$_{ans}$ & Acc$_{eq}$ & Acc$_{ans}$ & Acc$_{eq}$ & Acc$_{ans}$ & Acc$_{eq}$ & Acc$_{ans}$ & Acc & Acc & Acc \\ \midrule
\multicolumn{1}{c|}{} & Language & \cellcolor[HTML]{CFE1F3}-27.71\% & \cellcolor[HTML]{D2E2F4}-25.84\% & \cellcolor[HTML]{C3D9F0}-35.37\% & \cellcolor[HTML]{C0D8EF}-37.08\% & \cellcolor[HTML]{CFE1F3}-27.78\% & \cellcolor[HTML]{D0E1F3}-27.27\% & \cellcolor[HTML]{E9F0FA}-11.43\% & \cellcolor[HTML]{EAF1FA}-10.76\% & \cellcolor[HTML]{D9E7F5}-21.49\% & \cellcolor[HTML]{CCDFF2}-29.26\% & \cellcolor[HTML]{B5D1EC}-43.90\% \\
\multicolumn{1}{c|}{} & Type & \cellcolor[HTML]{A2C6E7}-55.42\% & \cellcolor[HTML]{E2ECF8}-15.73\% & \cellcolor[HTML]{ADCCEA}-48.78\% & \cellcolor[HTML]{D9E7F6}-21.35\% & \cellcolor[HTML]{BDD6EE}-38.89\% & \cellcolor[HTML]{D5E5F5}-23.64\% & \cellcolor[HTML]{FCFCFF}0.41\% & \cellcolor[HTML]{FCFCFF}1.59\% & \cellcolor[HTML]{E2ECF8}-15.59\% & \cellcolor[HTML]{D8E6F5}-21.96\% & \cellcolor[HTML]{E7EFF9}-12.60\% \\
\multicolumn{1}{c|}{} & Noise & \cellcolor[HTML]{A4C7E8}-54.22\% & \cellcolor[HTML]{AACAE9}-50.56\% & \cellcolor[HTML]{CEE0F3}-28.05\% & \cellcolor[HTML]{CCDFF2}-29.21\% & \cellcolor[HTML]{D8E6F5}-22.22\% & \cellcolor[HTML]{DBE8F6}-20.00\% & \cellcolor[HTML]{FCFCFF}0.41\% & \cellcolor[HTML]{FCFCFF}0.80\% & - & - & - \\
\multicolumn{1}{c|}{} & Distribution & \cellcolor[HTML]{D5E4F4}-24.10\% & \cellcolor[HTML]{DBE8F6}-20.22\% & \cellcolor[HTML]{B5D1EC}-43.90\% & \cellcolor[HTML]{D4E4F4}-24.72\% & \cellcolor[HTML]{CCDFF2}-29.63\% & \cellcolor[HTML]{CDDFF2}-29.09\% & \cellcolor[HTML]{ECF2FA}-9.80\% & \cellcolor[HTML]{ECF2FA}-9.56\% & - & - & - \\
\multicolumn{1}{c|}{} & Verbosity & \cellcolor[HTML]{99C0E5}-61.45\% & \cellcolor[HTML]{9AC1E5}-60.67\% & \cellcolor[HTML]{BFD7EF}-37.80\% & \cellcolor[HTML]{BAD4EE}-40.45\% & \cellcolor[HTML]{D5E4F4}-24.07\% & \cellcolor[HTML]{D3E3F4}-25.45\% & \cellcolor[HTML]{ABCBEA}-49.80\% & \cellcolor[HTML]{ACCCEA}-49.40\% & \cellcolor[HTML]{DCE9F6}-19.38\% & \cellcolor[HTML]{D1E2F3}-26.44\% & \cellcolor[HTML]{F4F7FD}-4.47\% \\
\multicolumn{1}{c|}{} & Extra & \cellcolor[HTML]{BFD7EF}-37.35\% & \cellcolor[HTML]{BED6EE}-38.20\% & \cellcolor[HTML]{97BFE4}-62.20\% & \cellcolor[HTML]{96BEE4}-62.92\% & \cellcolor[HTML]{93BDE3}-64.81\% & \cellcolor[HTML]{95BEE4}-63.64\% & \cellcolor[HTML]{7AAEDD}-80.41\% & \cellcolor[HTML]{7BAEDD}-79.68\% & \cellcolor[HTML]{D5E4F4}-23.86\% & \cellcolor[HTML]{D6E5F5}-23.17\% & \cellcolor[HTML]{F7F9FD}-2.85\% \\
\multicolumn{1}{c|}{} & Logic & \cellcolor[HTML]{B6D1EC}-43.37\% & \cellcolor[HTML]{BED6EE}-38.20\% & \cellcolor[HTML]{A3C6E7}-54.88\% & \cellcolor[HTML]{AECDEA}-48.31\% & \cellcolor[HTML]{9FC4E6}-57.41\% & \cellcolor[HTML]{A7C8E8}-52.73\% & \cellcolor[HTML]{B7D2ED}-42.45\% & \cellcolor[HTML]{C8DDF1}-31.87\% & - & - & \cellcolor[HTML]{C5DBF0}-33.69\% \\
\multicolumn{1}{c|}{\multirow{-8}{*}{Attack ($\Delta$\%)}} & Order & \cellcolor[HTML]{AACAE9}-50.60\% & \cellcolor[HTML]{EFF4FB}-7.87\% & \cellcolor[HTML]{ABCBE9}-50.00\% & \cellcolor[HTML]{F2F6FC}-5.62\% & \cellcolor[HTML]{96BEE4}-62.96\% & \cellcolor[HTML]{E1ECF8}-16.36\% & \cellcolor[HTML]{ABCBEA}-49.80\% & \cellcolor[HTML]{EBF1FA}-10.36\% & - & - & \cellcolor[HTML]{FCFCFF}2.64\% \\ \midrule
\multicolumn{1}{c|}{} & Language & \cellcolor[HTML]{DEEAF7}-18.07\% & \cellcolor[HTML]{DDE9F6}-19.10\% & \cellcolor[HTML]{CCDFF2}-29.27\% & \cellcolor[HTML]{C5DBF0}-33.71\% & \cellcolor[HTML]{FCFCFF}5.56\% & \cellcolor[HTML]{FCFCFF}5.45\% & \cellcolor[HTML]{E9F0FA}-11.43\% & \cellcolor[HTML]{EAF1FA}-10.76\% & \cellcolor[HTML]{FCFCFF}0.14\% & \cellcolor[HTML]{F6F8FD}-3.65\% & \cellcolor[HTML]{DFEAF7}-17.89\% \\
\multicolumn{1}{c|}{} & Type & \cellcolor[HTML]{E0EBF7}-16.87\% & \cellcolor[HTML]{E0EBF7}-16.85\% & \cellcolor[HTML]{F0F4FB}-7.32\% & \cellcolor[HTML]{EDF3FB}-8.99\% & \cellcolor[HTML]{FCFCFF}7.41\% & \cellcolor[HTML]{FCFCFF}9.09\% & \cellcolor[HTML]{FCFCFF}2.45\% & \cellcolor[HTML]{FCFCFF}2.79\% & \cellcolor[HTML]{FCFCFF}0.93\% & \cellcolor[HTML]{F8FAFE}-1.88\% & \cellcolor[HTML]{FCFCFF}6.91\% \\
\multicolumn{1}{c|}{} & Noise & \cellcolor[HTML]{D9E6F5}-21.69\% & \cellcolor[HTML]{DBE8F6}-20.22\% & \cellcolor[HTML]{F4F7FC}-4.88\% & \cellcolor[HTML]{F1F5FC}-6.74\% & \cellcolor[HTML]{FCFCFF}7.41\% & \cellcolor[HTML]{FCFCFF}9.09\% & \cellcolor[HTML]{FCFCFF}0.82\% & \cellcolor[HTML]{FCFCFF}0.40\% & - & - & - \\
\multicolumn{1}{c|}{} & Distribution & \cellcolor[HTML]{CBDEF2}-30.12\% & \cellcolor[HTML]{CEE0F3}-28.09\% & \cellcolor[HTML]{E8F0F9}-12.20\% & \cellcolor[HTML]{E6EEF9}-13.48\% & \cellcolor[HTML]{DEEAF7}-18.52\% & \cellcolor[HTML]{DBE8F6}-20.00\% & \cellcolor[HTML]{EBF2FA}-10.20\% & \cellcolor[HTML]{EDF3FB}-8.76\% & - & - & - \\
\multicolumn{1}{c|}{} & Verbosity & \cellcolor[HTML]{D7E5F5}-22.89\% & \cellcolor[HTML]{D7E6F5}-22.47\% & \cellcolor[HTML]{EEF3FB}-8.54\% & \cellcolor[HTML]{EBF2FA}-10.11\% & \cellcolor[HTML]{F9FAFE}-1.85\% & \cellcolor[HTML]{FBFBFE}0.00\% & \cellcolor[HTML]{FBFBFE}-0.41\% & \cellcolor[HTML]{FCFCFF}1.59\% & \cellcolor[HTML]{EBF1FA}-10.38\% & \cellcolor[HTML]{F6F8FD}-3.65\% & \cellcolor[HTML]{FCFCFF}5.31\% \\
\multicolumn{1}{c|}{} & Extra & \cellcolor[HTML]{FCFCFF}0.00\% & \cellcolor[HTML]{FCFCFF}2.25\% & \cellcolor[HTML]{F6F8FD}-3.66\% & \cellcolor[HTML]{F2F6FC}-5.62\% & \cellcolor[HTML]{BDD6EE}-38.89\% & \cellcolor[HTML]{BBD5EE}-40.00\% & \cellcolor[HTML]{CBDEF2}-30.20\% & \cellcolor[HTML]{D2E2F4}-25.90\% & \cellcolor[HTML]{D7E5F5}-22.90\% & \cellcolor[HTML]{DAE7F6}-20.72\% & \cellcolor[HTML]{D9E7F5}-21.54\% \\
\multicolumn{1}{c|}{} & Logic & - & - & - & - & - & - & - & - & - & - & 32.16\% \\
\multicolumn{1}{c|}{\multirow{-8}{*}{Defense ($\Delta$\%)}} & Order & \cellcolor[HTML]{BFD7EF}-37.35\% & \cellcolor[HTML]{F2F6FC}-5.62\% & \cellcolor[HTML]{B9D3ED}-41.46\% & \cellcolor[HTML]{EBF2FA}-10.11\% & \cellcolor[HTML]{F6F8FD}-3.70\% & \cellcolor[HTML]{FCFCFF}52.73\% & \cellcolor[HTML]{B5D1EC}-43.67\% & \cellcolor[HTML]{E9F0FA}-11.55\% & - & - & \cellcolor[HTML]{FCFCFF}45.92\% \\ \midrule
\multicolumn{1}{c|}{Original} & None & 68.03\% & 72.95\% & 67.21\% & 72.95\% & 44.26\% & 45.08\% & 66.94\% & 68.58\% & 49.42\% & 50.36\% & 42.41\% \\ \bottomrule
\end{tabular}%
}
\caption{The relative drop results of the DNC Framework. Notations here follow the ones in the main experiment results}
\label{tab:exp_relative}
\end{table*}

%% file: body/table_main_exp_with_coord.tex
\begin{table*}[ht]
\centering
\resizebox{\textwidth}{!}{%
\begin{tabular}{lccccccccccccc}
 &  &  & A & B & C & D & E & F & G & H & I & J & K \\ \cline{2-14} 
\multicolumn{1}{l|}{} & \multicolumn{2}{c|}{} & \multicolumn{8}{c|}{ASDiv-a} & \multicolumn{2}{c|}{DROP-num} & \multicolumn{1}{c|}{TATQA-a} \\ \cline{4-14} 
\multicolumn{1}{l|}{} & \multicolumn{2}{c|}{\multirow{-2}{*}{Configuration}} & \multicolumn{2}{c}{T5} & \multicolumn{2}{c}{BART} & \multicolumn{2}{c}{GPT2} & \multicolumn{2}{c|}{Graph2Tree} & T5 & \multicolumn{1}{c|}{BART} & \multicolumn{1}{c|}{TagOps} \\ \cline{2-14} 
\multicolumn{1}{l|}{} & Setting & \multicolumn{1}{c|}{Perturbation} & Acc$_{eq}$ & Acc$_{ans}$ & Acc$_{eq}$ & Acc$_{ans}$ & Acc$_{eq}$ & Acc$_{ans}$ & Acc$_{eq}$ & \multicolumn{1}{c|}{Acc$_{ans}$} & Acc & \multicolumn{1}{c|}{Acc} & \multicolumn{1}{c|}{Acc} \\ \cline{2-14} 
\multicolumn{1}{l|}{1} &  & \multicolumn{1}{c|}{Language} & \cellcolor[HTML]{BFD7EF}-18.85\% & \cellcolor[HTML]{BFD7EF}-18.85\% & \cellcolor[HTML]{AFCDEB}-23.77\% & \cellcolor[HTML]{A4C7E8}-27.05\% & \cellcolor[HTML]{D4E4F4}-12.30\% & \cellcolor[HTML]{D4E4F4}-12.30\% & \cellcolor[HTML]{E3EDF8}-7.65\% & \multicolumn{1}{c|}{\cellcolor[HTML]{E4EDF8}-7.38\%} & \cellcolor[HTML]{D9E7F6}-10.62\% & \multicolumn{1}{c|}{\cellcolor[HTML]{CCDFF2}-14.73\%} & \multicolumn{1}{c|}{\cellcolor[HTML]{C0D7EF}-18.62\%} \\
\multicolumn{1}{l|}{2} &  & \multicolumn{1}{c|}{Type} & \cellcolor[HTML]{82B2DF}-37.70\% & \cellcolor[HTML]{D7E5F5}-11.48\% & \cellcolor[HTML]{92BCE3}-32.79\% & \cellcolor[HTML]{C9DDF1}-15.57\% & \cellcolor[HTML]{C4DAF0}-17.21\% & \cellcolor[HTML]{D9E7F6}-10.66\% & \cellcolor[HTML]{FCFCFF}0.27\% & \multicolumn{1}{c|}{\cellcolor[HTML]{FCFCFF}1.09\%} & \cellcolor[HTML]{E3EDF8}-7.70\% & \multicolumn{1}{c|}{\cellcolor[HTML]{D8E6F5}-11.06\%} & \multicolumn{1}{c|}{\cellcolor[HTML]{EAF1FA}-5.34\%} \\
\multicolumn{1}{l|}{3} &  & \multicolumn{1}{c|}{Noise} & \cellcolor[HTML]{85B4E0}-36.89\% & \cellcolor[HTML]{85B4E0}-36.89\% & \cellcolor[HTML]{BFD7EF}-18.85\% & \cellcolor[HTML]{B7D2ED}-21.31\% & \cellcolor[HTML]{DCE8F6}-9.84\% & \cellcolor[HTML]{DEEAF7}-9.02\% & \cellcolor[HTML]{FCFCFF}0.27\% & \multicolumn{1}{c|}{\cellcolor[HTML]{FCFCFF}0.55\%} & - & \multicolumn{1}{c|}{-} & \multicolumn{1}{c|}{-} \\
\multicolumn{1}{l|}{4} &  & \multicolumn{1}{c|}{Distribution} & \cellcolor[HTML]{C7DCF1}-16.39\% & \cellcolor[HTML]{CCDFF2}-14.75\% & \cellcolor[HTML]{9CC2E6}-29.51\% & \cellcolor[HTML]{C1D9EF}-18.03\% & \cellcolor[HTML]{D1E2F3}-13.11\% & \cellcolor[HTML]{D1E2F3}-13.11\% & \cellcolor[HTML]{E6EFF9}-6.56\% & \multicolumn{1}{c|}{\cellcolor[HTML]{E6EFF9}-6.56\%} & - & \multicolumn{1}{c|}{-} & \multicolumn{1}{c|}{-} \\
\multicolumn{1}{l|}{5} &  & \multicolumn{1}{c|}{Verbosity} & \cellcolor[HTML]{75AADB}-41.80\% & \cellcolor[HTML]{6DA6D9}-44.26\% & \cellcolor[HTML]{AACAE9}-25.41\% & \cellcolor[HTML]{9CC2E6}-29.51\% & \cellcolor[HTML]{D9E7F6}-10.66\% & \cellcolor[HTML]{D7E5F5}-11.48\% & \cellcolor[HTML]{90BBE2}-33.33\% & \multicolumn{1}{c|}{\cellcolor[HTML]{8EBAE2}-33.88\%} & \cellcolor[HTML]{DDE9F6}-9.58\% & \multicolumn{1}{c|}{\cellcolor[HTML]{D1E2F3}-13.31\%} & \multicolumn{1}{c|}{\cellcolor[HTML]{F5F8FD}-1.90\%} \\
\multicolumn{1}{l|}{6} &  & \multicolumn{1}{c|}{Extra} & \cellcolor[HTML]{AACAE9}-25.41\% & \cellcolor[HTML]{A2C5E7}-27.87\% & \cellcolor[HTML]{75AADB}-41.80\% & \cellcolor[HTML]{68A2D8}-45.90\% & \cellcolor[HTML]{9FC4E6}-28.69\% & \cellcolor[HTML]{9FC4E6}-28.69\% & \cellcolor[HTML]{5B9BD5}-53.83\% & \multicolumn{1}{c|}{\cellcolor[HTML]{5B9BD5}-54.64\%} & \cellcolor[HTML]{D6E5F5}-11.79\% & \multicolumn{1}{c|}{\cellcolor[HTML]{D6E5F5}-11.67\%} & \multicolumn{1}{c|}{\cellcolor[HTML]{F8F9FD}-1.21\%} \\
\multicolumn{1}{l|}{7} &  & \multicolumn{1}{c|}{Logic} & \cellcolor[HTML]{9CC2E6}-29.51\% & \cellcolor[HTML]{A2C5E7}-27.87\% & \cellcolor[HTML]{85B4E0}-36.89\% & \cellcolor[HTML]{8AB7E1}-35.25\% & \cellcolor[HTML]{AACAE9}-25.41\% & \cellcolor[HTML]{AFCDEB}-23.77\% & \cellcolor[HTML]{A0C4E7}-28.42\% & \multicolumn{1}{c|}{\cellcolor[HTML]{B5D1EC}-21.86\%} & - & \multicolumn{1}{c|}{-} & \multicolumn{1}{c|}{\cellcolor[HTML]{CDE0F2}-14.29\%} \\
\multicolumn{1}{l|}{8} & \multirow{-8}{*}{Attack ($\Delta$)} & \multicolumn{1}{c|}{Order} & \cellcolor[HTML]{8DB9E2}-34.43\% & \cellcolor[HTML]{E9F0FA}-5.74\% & \cellcolor[HTML]{8FBAE2}-33.61\% & \cellcolor[HTML]{EEF4FB}-4.10\% & \cellcolor[HTML]{A2C5E7}-27.87\% & \cellcolor[HTML]{E4EDF8}-7.38\% & \cellcolor[HTML]{90BBE2}-33.33\% & \multicolumn{1}{c|}{\cellcolor[HTML]{E5EEF9}-7.10\%} & - & \multicolumn{1}{c|}{-} & \multicolumn{1}{c|}{\cellcolor[HTML]{FCFCFF}1.12\%} \\ \cline{2-14} 
\multicolumn{1}{l|}{9} &  & \multicolumn{1}{c|}{Language} & \cellcolor[HTML]{D4E4F4}-12.30\% & \cellcolor[HTML]{CFE0F3}-13.93\% & \cellcolor[HTML]{BCD5EE}-19.67\% & \cellcolor[HTML]{ACCCEA}-24.59\% & \cellcolor[HTML]{FCFCFF}2.46\% & \cellcolor[HTML]{FCFCFF}2.46\% & \cellcolor[HTML]{E3EDF8}-7.65\% & \multicolumn{1}{c|}{\cellcolor[HTML]{E4EDF8}-7.38\%} & \cellcolor[HTML]{FCFCFF}0.07\% & \multicolumn{1}{c|}{\cellcolor[HTML]{F6F8FD}-1.84\%} & \multicolumn{1}{c|}{\cellcolor[HTML]{E3EDF8}-7.59\%} \\
\multicolumn{1}{l|}{10} &  & \multicolumn{1}{c|}{Type} & \cellcolor[HTML]{D7E5F5}-11.48\% & \cellcolor[HTML]{D4E4F4}-12.30\% & \cellcolor[HTML]{ECF2FA}-4.92\% & \cellcolor[HTML]{E6EFF9}-6.56\% & \cellcolor[HTML]{FCFCFF}3.28\% & \cellcolor[HTML]{FCFCFF}4.10\% & \cellcolor[HTML]{FCFCFF}1.64\% & \multicolumn{1}{c|}{\cellcolor[HTML]{FCFCFF}1.91\%} & \cellcolor[HTML]{FCFCFF}0.46\% & \multicolumn{1}{c|}{\cellcolor[HTML]{F8FAFE}-0.95\%} & \multicolumn{1}{c|}{\cellcolor[HTML]{FCFCFF}2.93\%} \\
\multicolumn{1}{l|}{11} &  & \multicolumn{1}{c|}{Noise} & \cellcolor[HTML]{CCDFF2}-14.75\% & \cellcolor[HTML]{CCDFF2}-14.75\% & \cellcolor[HTML]{F1F5FC}-3.28\% & \cellcolor[HTML]{ECF2FA}-4.92\% & \cellcolor[HTML]{FCFCFF}3.28\% & \cellcolor[HTML]{FCFCFF}4.10\% & \cellcolor[HTML]{FCFCFF}0.55\% & \multicolumn{1}{c|}{\cellcolor[HTML]{FCFCFF}0.27\%} & - & \multicolumn{1}{c|}{-} & \multicolumn{1}{c|}{-} \\
\multicolumn{1}{l|}{12} &  & \multicolumn{1}{c|}{Distribution} & \cellcolor[HTML]{BAD4ED}-20.49\% & \cellcolor[HTML]{BAD4ED}-20.49\% & \cellcolor[HTML]{E1ECF8}-8.20\% & \cellcolor[HTML]{DCE8F6}-9.84\% & \cellcolor[HTML]{E1ECF8}-8.20\% & \cellcolor[HTML]{DEEAF7}-9.02\% & \cellcolor[HTML]{E6EEF9}-6.83\% & \multicolumn{1}{c|}{\cellcolor[HTML]{E8F0F9}-6.01\%} & - & \multicolumn{1}{c|}{-} & \multicolumn{1}{c|}{-} \\
\multicolumn{1}{l|}{13} &  & \multicolumn{1}{c|}{Verbosity} & \cellcolor[HTML]{C9DDF1}-15.57\% & \cellcolor[HTML]{C7DCF1}-16.39\% & \cellcolor[HTML]{E9F0FA}-5.74\% & \cellcolor[HTML]{E4EDF8}-7.38\% & \cellcolor[HTML]{F9FAFE}-0.82\% & \cellcolor[HTML]{FBFBFE}0.00\% & \cellcolor[HTML]{FBFBFE}-0.27\% & \multicolumn{1}{c|}{\cellcolor[HTML]{FCFCFF}1.09\%} & \cellcolor[HTML]{EBF2FA}-5.13\% & \multicolumn{1}{c|}{\cellcolor[HTML]{F6F8FD}-1.84\%} & \multicolumn{1}{c|}{\cellcolor[HTML]{FCFCFF}2.25\%} \\
\multicolumn{1}{l|}{14} &  & \multicolumn{1}{c|}{Extra} & \cellcolor[HTML]{FCFCFF}0.00\% & \cellcolor[HTML]{FCFCFF}1.64\% & \cellcolor[HTML]{F4F7FC}-2.46\% & \cellcolor[HTML]{EEF4FB}-4.10\% & \cellcolor[HTML]{C4DAF0}-17.21\% & \cellcolor[HTML]{C1D9EF}-18.03\% & \cellcolor[HTML]{BAD4EE}-20.22\% & \multicolumn{1}{c|}{\cellcolor[HTML]{C2D9F0}-17.76\%} & \cellcolor[HTML]{D7E6F5}-11.32\% & \multicolumn{1}{c|}{\cellcolor[HTML]{DAE7F6}-10.44\%} & \multicolumn{1}{c|}{\cellcolor[HTML]{DEEAF7}-9.14\%} \\
\multicolumn{1}{l|}{15} &  & \multicolumn{1}{c|}{Logic} & - & - & - & - & - & - & - & \multicolumn{1}{c|}{-} & - & \multicolumn{1}{c|}{-} & \multicolumn{1}{c|}{\cellcolor[HTML]{FCFCFF}13.64\%} \\
\multicolumn{1}{l|}{16} & \multirow{-8}{*}{Defense ($\Delta$)} & \multicolumn{1}{c|}{Order} & \cellcolor[HTML]{AACAE9}-25.41\% & \cellcolor[HTML]{EEF4FB}-4.10\% & \cellcolor[HTML]{A2C5E7}-27.87\% & \cellcolor[HTML]{E4EDF8}-7.38\% & \cellcolor[HTML]{F6F8FD}-1.64\% & \cellcolor[HTML]{FCFCFF}23.77\% & \cellcolor[HTML]{9DC3E6}-29.23\% & \multicolumn{1}{c|}{\cellcolor[HTML]{E2ECF8}-7.92\%} & - & \multicolumn{1}{c|}{-} & \multicolumn{1}{c|}{\cellcolor[HTML]{FCFCFF}19.47\%} \\ \cline{2-14} 
\multicolumn{1}{l|}{} & Original & \multicolumn{1}{c|}{None} & 68.03\% & 72.95\% & 67.21\% & 72.95\% & 44.26\% & 45.08\% & 66.94\% & \multicolumn{1}{c|}{68.58\%} & 49.42\% & \multicolumn{1}{c|}{50.36\%} & \multicolumn{1}{c|}{42.41\%} \\ \cline{2-14} 
\end{tabular}%
}
\caption{The main experiment result table with cell coordinates}
\label{tab:main_exp_with_coord}
\end{table*}

\begin{table*}[ht]
\centering
\resizebox{\textwidth}{!}{%
\begin{tabular}{@{}clc@{}}
\toprule
\textbf{Value} & \multicolumn{1}{c}{\textbf{Explanation}} & \textbf{Formula} \\ \midrule
19.66\% & \begin{tabular}[c]{@{}l@{}}Average performance drop of all models \\ caused by perturbations according to the \\ Semantic Parsing stage\end{tabular} & AVERAGE(B5:B8,D5:D8,F5:F8,H5:H8,I5:I8,J5:J8,K5:K8) \\ \midrule
13.15\% & \begin{tabular}[c]{@{}l@{}}Average performance drop of all models\\ caused by perturbations according to the\\ Numerical Parsing stage\end{tabular} & AVERAGE(B1:B4,D1:D4,F1:F4,H1:H4,I1:I4,J1:J4,K1:K4) \\ \midrule
17.42\% & \begin{tabular}[c]{@{}l@{}}Average performance drop of all Transformer-\\ based models caused by perturbations according \\ to the Semantic Parsing stage\end{tabular} & AVERAGE(B5:B8,D5:D8,F5:F8,I5:I8,J5:J8,K5:K8) \\ \midrule
3.07\% & \begin{tabular}[c]{@{}l@{}}Average performance drop of the Graph2Tree\\ system caused by perturbations   according to\\ the Semantic Parsing stage\end{tabular} & AVERAGE(B1:B4,D1:D4,F1:F4,I1:I4,J1:J4,K1:K4) \\ \midrule
17.96\% & \begin{tabular}[c]{@{}l@{}}Average performance recovery of all models\\ caused by perturbations according to the\\ Semantic Parsing stage\end{tabular} & AVERAGE(B13:B16,D13:D16,F13:F16,H13:H16,I13:I16,J13:J16,K13:K16) \\ \midrule
6.52\% & \begin{tabular}[c]{@{}l@{}}Average performance recovery of all models\\ caused by perturbations according to the\\ Numerical Parsing stage\end{tabular} & AVERAGE(B9:B12,D9:D12,F9:F12,H9:H12,I9:I12,J9:J12,K9:K12) \\ \midrule
12.53\% & \begin{tabular}[c]{@{}l@{}}Average performance recovery of all Transformer-\\ based models caused by perturbations according\\ to the Semantic Parsing stage\end{tabular} & AVERAGE(B13:B16,D13:D16,F13:F16,I13:I16,J13:J16,K13:K16) \\ \midrule
11.58\% & \begin{tabular}[c]{@{}l@{}}Average performance recovery of the Graph2Tree\\ system caused by perturbations according to the\\ Semantic Parsing stage\end{tabular} & AVERAGE(B9:B12,D9:D12,F9:F12,I9:I12,J9:J12,K9:K12) \\ \bottomrule
\end{tabular}%
}
\caption{The Value, Explanation, And Formula Used of The Experiment Observations.}
\label{tab:exp_formula}
\end{table*}

%% file: main.bbl
\begin{thebibliography}{46}
\expandafter\ifx\csname natexlab\endcsname\relax\def\natexlab#1{#1}\fi

\bibitem[{Al-Negheimish et~al.(2021{\natexlab{a}})Al-Negheimish, Madhyastha,
  and Russo}]{al-negheimish-etal-2021-discrete}
Hadeel Al-Negheimish, Pranava Madhyastha, and Alessandra Russo.
  2021{\natexlab{a}}.
\newblock \href {https://doi.org/10.18653/v1/2021.eacl-srw.12} {Discrete
  reasoning templates for natural language understanding}.
\newblock In \emph{Proceedings of the 16th Conference of the European Chapter
  of the Association for Computational Linguistics: Student Research Workshop},
  pages 80--87, Online. Association for Computational Linguistics.

\bibitem[{Al-Negheimish et~al.(2021{\natexlab{b}})Al-Negheimish, Madhyastha,
  and Russo}]{al-negheimish-etal-2021-numerical}
Hadeel Al-Negheimish, Pranava Madhyastha, and Alessandra Russo.
  2021{\natexlab{b}}.
\newblock \href {https://doi.org/10.18653/v1/2021.emnlp-main.759} {Numerical
  reasoning in machine reading comprehension tasks: are we there yet?}
\newblock In \emph{Proceedings of the 2021 Conference on Empirical Methods in
  Natural Language Processing}, pages 9643--9649, Online and Punta Cana,
  Dominican Republic. Association for Computational Linguistics.

\bibitem[{Chen et~al.(2020{\natexlab{a}})Chen, Xu, Cheng, Xiaochuan, Zhang,
  Song, Wang, Qi, and Chu}]{chen-etal-2020-question}
Kunlong Chen, Weidi Xu, Xingyi Cheng, Zou Xiaochuan, Yuyu Zhang, Le~Song,
  Taifeng Wang, Yuan Qi, and Wei Chu. 2020{\natexlab{a}}.
\newblock \href {https://doi.org/10.18653/v1/2020.emnlp-main.549} {Question
  directed graph attention network for numerical reasoning over text}.
\newblock In \emph{Proceedings of the 2020 Conference on Empirical Methods in
  Natural Language Processing (EMNLP)}, pages 6759--6768, Online. Association
  for Computational Linguistics.

\bibitem[{Chen et~al.(2020{\natexlab{b}})Chen, Wang, Chen, Zhang, Wang, Li,
  Zhou, and Wang}]{2019TabFactA}
Wenhu Chen, Hongmin Wang, Jianshu Chen, Yunkai Zhang, Hong Wang, Shiyang Li,
  Xiyou Zhou, and William~Yang Wang. 2020{\natexlab{b}}.
\newblock Tabfact : A large-scale dataset for table-based fact verification.
\newblock In \emph{International Conference on Learning Representations
  (ICLR)}, Addis Ababa, Ethiopia.

\bibitem[{Chen et~al.(2021)Chen, Chen, Smiley, Shah, Borova, Langdon, Moussa,
  Beane, Huang, Routledge, and Wang}]{chen-etal-2021-finqa}
Zhiyu Chen, Wenhu Chen, Charese Smiley, Sameena Shah, Iana Borova, Dylan
  Langdon, Reema Moussa, Matt Beane, Ting-Hao Huang, Bryan Routledge, and
  William~Yang Wang. 2021.
\newblock \href {https://doi.org/10.18653/v1/2021.emnlp-main.300} {{F}in{QA}: A
  dataset of numerical reasoning over financial data}.
\newblock In \emph{Proceedings of the 2021 Conference on Empirical Methods in
  Natural Language Processing}, pages 3697--3711, Online and Punta Cana,
  Dominican Republic. Association for Computational Linguistics.

\bibitem[{Chiang and Chen(2019)}]{chiang-chen-2019-semantically}
Ting-Rui Chiang and Yun-Nung Chen. 2019.
\newblock \href {https://doi.org/10.18653/v1/N19-1272} {Semantically-aligned
  equation generation for solving and reasoning math word problems}.
\newblock In \emph{Proceedings of the 2019 Conference of the North {A}merican
  Chapter of the Association for Computational Linguistics: Human Language
  Technologies, Volume 1 (Long and Short Papers)}, pages 2656--2668,
  Minneapolis, Minnesota. Association for Computational Linguistics.

\bibitem[{Chowdhery et~al.(2022)Chowdhery, Narang, Devlin, Bosma, Mishra,
  Roberts, Barham, Chung, Sutton, Gehrmann, Schuh, Shi, Tsvyashchenko, Maynez,
  Rao, Barnes, Tay, Shazeer, Prabhakaran, Reif, Du, Hutchinson, Pope, Bradbury,
  Austin, Isard, Gur-Ari, Yin, Duke, Levskaya, Ghemawat, Dev, Michalewski,
  Garc{\'i}a, Misra, Robinson, Fedus, Zhou, Ippolito, Luan, Lim, Zoph,
  Spiridonov, Sepassi, Dohan, Agrawal, Omernick, Dai, Pillai, Pellat,
  Lewkowycz, Moreira, Child, Polozov, Lee, Zhou, Wang, Saeta, D{\'i}az, Firat,
  Catasta, Wei, Meier-Hellstern, Eck, Dean, Petrov, and
  Fiedel}]{Chowdhery2022PaLMSL}
Aakanksha Chowdhery, Sharan Narang, Jacob Devlin, Maarten Bosma, Gaurav Mishra,
  Adam Roberts, Paul Barham, Hyung~Won Chung, Charles Sutton, Sebastian
  Gehrmann, Parker Schuh, Kensen Shi, Sasha Tsvyashchenko, Joshua Maynez,
  Abhishek~B Rao, Parker Barnes, Yi~Tay, Noam~M. Shazeer, Vinodkumar
  Prabhakaran, Emily Reif, Nan Du, Benton~C. Hutchinson, Reiner Pope, James
  Bradbury, Jacob Austin, Michael Isard, Guy Gur-Ari, Pengcheng Yin, Toju Duke,
  Anselm Levskaya, Sanjay Ghemawat, Sunipa Dev, Henryk Michalewski, Xavier
  Garc{\'i}a, Vedant Misra, Kevin Robinson, Liam Fedus, Denny Zhou, Daphne
  Ippolito, David Luan, Hyeontaek Lim, Barret Zoph, Alexander Spiridonov, Ryan
  Sepassi, David Dohan, Shivani Agrawal, Mark Omernick, Andrew~M. Dai,
  Thanumalayan~Sankaranarayana Pillai, Marie Pellat, Aitor Lewkowycz,
  Erica~Oliveira Moreira, Rewon Child, Oleksandr Polozov, Katherine Lee,
  Zongwei Zhou, Xuezhi Wang, Brennan Saeta, Mark D{\'i}az, Orhan Firat, Michele
  Catasta, Jason Wei, Kathleen~S. Meier-Hellstern, Douglas Eck, Jeff Dean, Slav
  Petrov, and Noah Fiedel. 2022.
\newblock Palm: Scaling language modeling with pathways.
\newblock \emph{ArXiv}, abs/2204.02311.

\bibitem[{Dong et~al.(2022)Dong, Cheng, He, Zhou, Zhou, Zhou, Liu, Han, and
  Zhang}]{dong2022table}
Haoyu Dong, Zhoujun Cheng, Xinyi He, Mengyu Zhou, Anda Zhou, Fan Zhou, Ao~Liu,
  Shi Han, and Dongmei Zhang. 2022.
\newblock \href
  {https://www.microsoft.com/en-us/research/publication/table-pre-training-a-survey-on-model-architectures-pre-training-objectives-and-downstream-tasks/}
  {Table pre-training: A survey on model architectures, pre-training
  objectives, and downstream tasks}.
\newblock In \emph{IJCAI'2022 SURVEY TRACK}.

\bibitem[{Dua et~al.(2019)Dua, Wang, Dasigi, Stanovsky, Singh, and
  Gardner}]{dua-etal-2019-drop}
Dheeru Dua, Yizhong Wang, Pradeep Dasigi, Gabriel Stanovsky, Sameer Singh, and
  Matt Gardner. 2019.
\newblock \href {https://doi.org/10.18653/v1/N19-1246} {{DROP}: A reading
  comprehension benchmark requiring discrete reasoning over paragraphs}.
\newblock In \emph{Proceedings of the 2019 Conference of the North {A}merican
  Chapter of the Association for Computational Linguistics: Human Language
  Technologies, Volume 1 (Long and Short Papers)}, pages 2368--2378,
  Minneapolis, Minnesota. Association for Computational Linguistics.

\bibitem[{Herzig et~al.(2020)Herzig, Nowak, M{\"u}ller, Piccinno, and
  Eisenschlos}]{herzig-etal-2020-tapas}
Jonathan Herzig, Pawel~Krzysztof Nowak, Thomas M{\"u}ller, Francesco Piccinno,
  and Julian Eisenschlos. 2020.
\newblock \href {https://doi.org/10.18653/v1/2020.acl-main.398} {{T}a{P}as:
  Weakly supervised table parsing via pre-training}.
\newblock In \emph{Proceedings of the 58th Annual Meeting of the Association
  for Computational Linguistics}, pages 4320--4333, Online. Association for
  Computational Linguistics.

\bibitem[{Hu et~al.(2019)Hu, Peng, Huang, and Li}]{hu-etal-2019-multi}
Minghao Hu, Yuxing Peng, Zhen Huang, and Dongsheng Li. 2019.
\newblock \href {https://doi.org/10.18653/v1/D19-1170} {A multi-type multi-span
  network for reading comprehension that requires discrete reasoning}.
\newblock In \emph{Proceedings of the 2019 Conference on Empirical Methods in
  Natural Language Processing and the 9th International Joint Conference on
  Natural Language Processing (EMNLP-IJCNLP)}, pages 1596--1606, Hong Kong,
  China. Association for Computational Linguistics.

\bibitem[{Iida et~al.(2021)Iida, Thai, Manjunatha, and
  Iyyer}]{iida-etal-2021-tabbie}
Hiroshi Iida, Dung Thai, Varun Manjunatha, and Mohit Iyyer. 2021.
\newblock \href {https://doi.org/10.18653/v1/2021.naacl-main.270} {{TABBIE}:
  Pretrained representations of tabular data}.
\newblock In \emph{Proceedings of the 2021 Conference of the North American
  Chapter of the Association for Computational Linguistics: Human Language
  Technologies}, pages 3446--3456, Online. Association for Computational
  Linguistics.

\bibitem[{Kim et~al.(2021)Kim, Hong, Kim, Kang, and
  Myaeng}]{kim-etal-2021-seen}
Jeonghwan Kim, Giwon Hong, Kyung-min Kim, Junmo Kang, and Sung-Hyon Myaeng.
  2021.
\newblock \href {https://doi.org/10.18653/v1/2021.emnlp-main.563} {Have you
  seen that number? investigating extrapolation in question answering models}.
\newblock In \emph{Proceedings of the 2021 Conference on Empirical Methods in
  Natural Language Processing}, pages 7031--7037, Online and Punta Cana,
  Dominican Republic. Association for Computational Linguistics.

\bibitem[{Koncel-Kedziorski et~al.(2016)Koncel-Kedziorski, Roy, Amini, Kushman,
  and Hajishirzi}]{koncel-kedziorski-etal-2016-mawps}
Rik Koncel-Kedziorski, Subhro Roy, Aida Amini, Nate Kushman, and Hannaneh
  Hajishirzi. 2016.
\newblock \href {https://doi.org/10.18653/v1/N16-1136} {{MAWPS}: A math word
  problem repository}.
\newblock In \emph{Proceedings of the 2016 Conference of the North {A}merican
  Chapter of the Association for Computational Linguistics: Human Language
  Technologies}, pages 1152--1157, San Diego, California. Association for
  Computational Linguistics.

\bibitem[{Kumar et~al.(2021)Kumar, Maheshwary, and
  Pudi}]{kumar-etal-2021-adversarial-examples}
Vivek Kumar, Rishabh Maheshwary, and Vikram Pudi. 2021.
\newblock \href {https://doi.org/10.18653/v1/2021.findings-emnlp.230}
  {Adversarial examples for evaluating math word problem solvers}.
\newblock In \emph{Findings of the Association for Computational Linguistics:
  EMNLP 2021}, pages 2705--2712, Punta Cana, Dominican Republic. Association
  for Computational Linguistics.

\bibitem[{Kushman et~al.(2014)Kushman, Artzi, Zettlemoyer, and
  Barzilay}]{kushman-etal-2014-learning}
Nate Kushman, Yoav Artzi, Luke Zettlemoyer, and Regina Barzilay. 2014.
\newblock \href {https://doi.org/10.3115/v1/P14-1026} {Learning to
  automatically solve algebra word problems}.
\newblock In \emph{Proceedings of the 52nd Annual Meeting of the Association
  for Computational Linguistics (Volume 1: Long Papers)}, pages 271--281,
  Baltimore, Maryland. Association for Computational Linguistics.

\bibitem[{Lan et~al.(2022)Lan, Wang, Zhang, Lan, Dai, Wang, Zhang, and
  Lim}]{Lan_Wang_Zhang_Lan_Dai_Wang_Zhang_Lim_2022}
Yihuai Lan, Lei Wang, Qiyuan Zhang, Yunshi Lan, Bing~Tian Dai, Yan Wang,
  Dongxiang Zhang, and Ee-Peng Lim. 2022.
\newblock \href {https://doi.org/10.1609/aaai.v36i11.21723} {Mwptoolkit: An
  open-source framework for deep learning-based math word problem solvers}.
\newblock \emph{Proceedings of the AAAI Conference on Artificial Intelligence},
  36(11):13188--13190.

\bibitem[{Liang et~al.(2021)Liang, Zhang, Shao, and
  Zhang}]{DBLP:journals/corr/abs-2107-13435}
Zhenwen Liang, Jipeng Zhang, Jie Shao, and Xiangliang Zhang. 2021.
\newblock \href {http://arxiv.org/abs/2107.13435} {{MWP-BERT:} {A} strong
  baseline for math word problems}.
\newblock \emph{CoRR}, abs/2107.13435.

\bibitem[{Liu et~al.(2022)Liu, Chen, Guo, Ziyadi, Lin, Chen, and
  Lou}]{liu2022tapex}
Qian Liu, Bei Chen, Jiaqi Guo, Morteza Ziyadi, Zeqi Lin, Weizhu Chen, and
  Jian-Guang Lou. 2022.
\newblock \href {https://openreview.net/forum?id=O50443AsCP} {{TAPEX}: Table
  pre-training via learning a neural {SQL} executor}.
\newblock In \emph{International Conference on Learning Representations}.

\bibitem[{Miao et~al.(2020)Miao, Liang, and Su}]{miao-etal-2020-diverse}
Shen-yun Miao, Chao-Chun Liang, and Keh-Yih Su. 2020.
\newblock \href {https://doi.org/10.18653/v1/2020.acl-main.92} {A diverse
  corpus for evaluating and developing {E}nglish math word problem solvers}.
\newblock In \emph{Proceedings of the 58th Annual Meeting of the Association
  for Computational Linguistics}, pages 975--984, Online. Association for
  Computational Linguistics.

\bibitem[{Nogueira et~al.(2021)Nogueira, Jiang, and
  Lin}]{DBLP:journals/corr/abs-2102-13019}
Rodrigo Nogueira, Zhiying Jiang, and Jimmy Lin. 2021.
\newblock \href {http://arxiv.org/abs/2102.13019} {Investigating the
  limitations of the transformers with simple arithmetic tasks}.
\newblock \emph{CoRR}, abs/2102.13019.

\bibitem[{Padhi et~al.(2021)Padhi, Schiff, Melnyk, Rigotti, Mroueh, Dognin,
  Ross, Nair, and Altman}]{padhi2021tabular}
Inkit Padhi, Yair Schiff, Igor Melnyk, Mattia Rigotti, Youssef Mroueh, Pierre
  Dognin, Jerret Ross, Ravi Nair, and Erik Altman. 2021.
\newblock \href {https://ieeexplore.ieee.org/document/9414142} {Tabular
  transformers for modeling multivariate time series}.
\newblock In \emph{ICASSP 2021-2021 IEEE International Conference on Acoustics,
  Speech and Signal Processing (ICASSP)}, pages 3565--3569. IEEE.

\bibitem[{Pal and Baral(2021)}]{pal-baral-2021-investigating-numeracy}
Kuntal~Kumar Pal and Chitta Baral. 2021.
\newblock \href {https://doi.org/10.18653/v1/2021.findings-emnlp.265}
  {Investigating numeracy learning ability of a text-to-text transfer model}.
\newblock In \emph{Findings of the Association for Computational Linguistics:
  EMNLP 2021}, pages 3095--3101, Punta Cana, Dominican Republic. Association
  for Computational Linguistics.

\bibitem[{Pasupat and Liang(2015)}]{pasupat-liang-2015-compositional}
Panupong Pasupat and Percy Liang. 2015.
\newblock \href {https://doi.org/10.3115/v1/P15-1142} {Compositional semantic
  parsing on semi-structured tables}.
\newblock In \emph{Proceedings of the 53rd Annual Meeting of the Association
  for Computational Linguistics and the 7th International Joint Conference on
  Natural Language Processing (Volume 1: Long Papers)}, pages 1470--1480,
  Beijing, China. Association for Computational Linguistics.

\bibitem[{Patel et~al.(2021)Patel, Bhattamishra, and
  Goyal}]{patel-etal-2021-nlp}
Arkil Patel, Satwik Bhattamishra, and Navin Goyal. 2021.
\newblock \href {https://doi.org/10.18653/v1/2021.naacl-main.168} {Are {NLP}
  models really able to solve simple math word problems?}
\newblock In \emph{Proceedings of the 2021 Conference of the North American
  Chapter of the Association for Computational Linguistics: Human Language
  Technologies}, pages 2080--2094, Online. Association for Computational
  Linguistics.

\bibitem[{Pi et~al.(2022{\natexlab{a}})Pi, Liu, Chen, Ziyadi, Lin, Gao, Fu,
  Lou, and Chen}]{Pi2022ReasoningLP}
Xinyu Pi, Qian Liu, Bei Chen, Morteza Ziyadi, Zeqi Lin, Yan Gao, Qiang Fu,
  Jian-Guang Lou, and Weizhu Chen. 2022{\natexlab{a}}.
\newblock Reasoning like program executors.
\newblock \emph{ArXiv}, abs/2201.11473.

\bibitem[{Pi et~al.(2022{\natexlab{b}})Pi, Wang, Gao, Guo, Li, and
  Lou}]{pi-etal-2022-towards}
Xinyu Pi, Bing Wang, Yan Gao, Jiaqi Guo, Zhoujun Li, and Jian-Guang Lou.
  2022{\natexlab{b}}.
\newblock \href {https://aclanthology.org/2022.acl-long.142} {Towards
  robustness of text-to-{SQL} models against natural and realistic adversarial
  table perturbation}.
\newblock In \emph{Proceedings of the 60th Annual Meeting of the Association
  for Computational Linguistics (Volume 1: Long Papers)}, pages 2007--2022,
  Dublin, Ireland. Association for Computational Linguistics.

\bibitem[{Qin et~al.(2020)Qin, Lin, Liang, Zhang, and
  Lin}]{qin-etal-2020-semantically}
Jinghui Qin, Lihui Lin, Xiaodan Liang, Rumin Zhang, and Liang Lin. 2020.
\newblock \href {https://doi.org/10.18653/v1/2020.emnlp-main.309}
  {Semantically-aligned universal tree-structured solver for math word
  problems}.
\newblock In \emph{Proceedings of the 2020 Conference on Empirical Methods in
  Natural Language Processing (EMNLP)}, pages 3780--3789, Online. Association
  for Computational Linguistics.

\bibitem[{Ran et~al.(2019)Ran, Lin, Li, Zhou, and Liu}]{ran-etal-2019-numnet}
Qiu Ran, Yankai Lin, Peng Li, Jie Zhou, and Zhiyuan Liu. 2019.
\newblock \href {https://doi.org/10.18653/v1/D19-1251} {{N}um{N}et: Machine
  reading comprehension with numerical reasoning}.
\newblock In \emph{Proceedings of the 2019 Conference on Empirical Methods in
  Natural Language Processing and the 9th International Joint Conference on
  Natural Language Processing (EMNLP-IJCNLP)}, pages 2474--2484, Hong Kong,
  China. Association for Computational Linguistics.

\bibitem[{Sennrich et~al.(2016)Sennrich, Haddow, and
  Birch}]{sennrich-etal-2016-neural}
Rico Sennrich, Barry Haddow, and Alexandra Birch. 2016.
\newblock \href {https://doi.org/10.18653/v1/P16-1162} {Neural machine
  translation of rare words with subword units}.
\newblock In \emph{Proceedings of the 54th Annual Meeting of the Association
  for Computational Linguistics (Volume 1: Long Papers)}, pages 1715--1725,
  Berlin, Germany. Association for Computational Linguistics.

\bibitem[{Shen and Jin(2020)}]{shen-jin-2020-solving}
Yibin Shen and Cheqing Jin. 2020.
\newblock \href {https://doi.org/10.18653/v1/2020.coling-main.262} {Solving
  math word problems with multi-encoders and multi-decoders}.
\newblock In \emph{Proceedings of the 28th International Conference on
  Computational Linguistics}, pages 2924--2934, Barcelona, Spain (Online).
  International Committee on Computational Linguistics.

\bibitem[{Spithourakis and Riedel(2018)}]{spithourakis-riedel-2018-numeracy}
Georgios Spithourakis and Sebastian Riedel. 2018.
\newblock \href {https://doi.org/10.18653/v1/P18-1196} {Numeracy for language
  models: Evaluating and improving their ability to predict numbers}.
\newblock In \emph{Proceedings of the 56th Annual Meeting of the Association
  for Computational Linguistics (Volume 1: Long Papers)}, pages 2104--2115,
  Melbourne, Australia. Association for Computational Linguistics.

\bibitem[{Sundaram et~al.(2022)Sundaram, Gurajada, Fisichella, P, and
  Abraham}]{Sundaram2022WhyAN}
Sowmya~S. Sundaram, Sairam Gurajada, Marco Fisichella, Deepak P, and
  Savitha~Sam Abraham. 2022.
\newblock Why are nlp models fumbling at elementary math? a survey of deep
  learning based word problem solvers.
\newblock \emph{ArXiv}, abs/2205.15683.

\bibitem[{Sundararaman et~al.(2020)Sundararaman, Si, Subramanian, Wang,
  Hazarika, and Carin}]{sundararaman-etal-2020-methods}
Dhanasekar Sundararaman, Shijing Si, Vivek Subramanian, Guoyin Wang, Devamanyu
  Hazarika, and Lawrence Carin. 2020.
\newblock \href {https://doi.org/10.18653/v1/2020.emnlp-main.384} {Methods for
  numeracy-preserving word embeddings}.
\newblock In \emph{Proceedings of the 2020 Conference on Empirical Methods in
  Natural Language Processing (EMNLP)}, pages 4742--4753, Online. Association
  for Computational Linguistics.

\bibitem[{Thawani et~al.(2021{\natexlab{a}})Thawani, Pujara, and
  Ilievski}]{thawani-etal-2021-numeracy}
Avijit Thawani, Jay Pujara, and Filip Ilievski. 2021{\natexlab{a}}.
\newblock \href {https://doi.org/10.18653/v1/2021.emnlp-main.557} {Numeracy
  enhances the literacy of language models}.
\newblock In \emph{Proceedings of the 2021 Conference on Empirical Methods in
  Natural Language Processing}, pages 6960--6967, Online and Punta Cana,
  Dominican Republic. Association for Computational Linguistics.

\bibitem[{Thawani et~al.(2021{\natexlab{b}})Thawani, Pujara, Ilievski, and
  Szekely}]{thawani-etal-2021-representing}
Avijit Thawani, Jay Pujara, Filip Ilievski, and Pedro Szekely.
  2021{\natexlab{b}}.
\newblock \href {https://doi.org/10.18653/v1/2021.naacl-main.53} {Representing
  numbers in {NLP}: a survey and a vision}.
\newblock In \emph{Proceedings of the 2021 Conference of the North American
  Chapter of the Association for Computational Linguistics: Human Language
  Technologies}, pages 644--656, Online. Association for Computational
  Linguistics.

\bibitem[{Upadhyay and Chang(2017)}]{upadhyay-chang-2017-annotating}
Shyam Upadhyay and Ming-Wei Chang. 2017.
\newblock \href {https://aclanthology.org/E17-1047} {Annotating derivations: A
  new evaluation strategy and dataset for algebra word problems}.
\newblock In \emph{Proceedings of the 15th Conference of the {E}uropean Chapter
  of the Association for Computational Linguistics: Volume 1, Long Papers},
  pages 494--504, Valencia, Spain. Association for Computational Linguistics.

\bibitem[{Wallace et~al.(2019)Wallace, Wang, Li, Singh, and
  Gardner}]{wallace-etal-2019-nlp}
Eric Wallace, Yizhong Wang, Sujian Li, Sameer Singh, and Matt Gardner. 2019.
\newblock \href {https://doi.org/10.18653/v1/D19-1534} {Do {NLP} models know
  numbers? probing numeracy in embeddings}.
\newblock In \emph{Proceedings of the 2019 Conference on Empirical Methods in
  Natural Language Processing and the 9th International Joint Conference on
  Natural Language Processing (EMNLP-IJCNLP)}, pages 5307--5315, Hong Kong,
  China. Association for Computational Linguistics.

\bibitem[{Wang et~al.(2017)Wang, Liu, and Shi}]{wang-etal-2017-deep}
Yan Wang, Xiaojiang Liu, and Shuming Shi. 2017.
\newblock \href {https://doi.org/10.18653/v1/D17-1088} {Deep neural solver for
  math word problems}.
\newblock In \emph{Proceedings of the 2017 Conference on Empirical Methods in
  Natural Language Processing}, pages 845--854, Copenhagen, Denmark.
  Association for Computational Linguistics.

\bibitem[{Wu et~al.(2016)Wu, Schuster, Chen, Le, Norouzi, Macherey, Krikun,
  Cao, Gao, Macherey, Klingner, Shah, Johnson, Liu, Kaiser, Gouws, Kato, Kudo,
  Kazawa, Stevens, Kurian, Patil, Wang, Young, Smith, Riesa, Rudnick, Vinyals,
  Corrado, Hughes, and Dean}]{Wu2016GooglesNM}
Yonghui Wu, Mike Schuster, Z.~Chen, Quoc~V. Le, Mohammad Norouzi, Wolfgang
  Macherey, Maxim Krikun, Yuan Cao, Qin Gao, Klaus Macherey, Jeff Klingner,
  Apurva Shah, Melvin Johnson, Xiaobing Liu, Lukasz Kaiser, Stephan Gouws,
  Yoshikiyo Kato, Taku Kudo, Hideto Kazawa, Keith Stevens, George Kurian,
  Nishant Patil, Wei Wang, Cliff Young, Jason~R. Smith, Jason Riesa, Alex
  Rudnick, Oriol Vinyals, Gregory~S. Corrado, Macduff Hughes, and Jeffrey Dean.
  2016.
\newblock Google's neural machine translation system: Bridging the gap between
  human and machine translation.
\newblock \emph{ArXiv}, abs/1609.08144.

\bibitem[{Xie and Sun(2019)}]{ijcai2019-736}
Zhipeng Xie and Shichao Sun. 2019.
\newblock \href {https://doi.org/10.24963/ijcai.2019/736} {A goal-driven
  tree-structured neural model for math word problems}.
\newblock In \emph{Proceedings of the Twenty-Eighth International Joint
  Conference on Artificial Intelligence, {IJCAI-19}}, pages 5299--5305.
  International Joint Conferences on Artificial Intelligence Organization.

\bibitem[{Yin et~al.(2020)Yin, Neubig, Yih, and Riedel}]{yin-etal-2020-tabert}
Pengcheng Yin, Graham Neubig, Wen-tau Yih, and Sebastian Riedel. 2020.
\newblock \href {https://doi.org/10.18653/v1/2020.acl-main.745} {{T}a{BERT}:
  Pretraining for joint understanding of textual and tabular data}.
\newblock In \emph{Proceedings of the 58th Annual Meeting of the Association
  for Computational Linguistics}, pages 8413--8426, Online. Association for
  Computational Linguistics.

\bibitem[{Zhang et~al.(2020{\natexlab{a}})Zhang, Wang, Zhang, Dai, and
  Shen}]{8703135}
Dongxiang Zhang, Lei Wang, Luming Zhang, Bing~Tian Dai, and Heng~Tao Shen.
  2020{\natexlab{a}}.
\newblock \href {https://doi.org/10.1109/TPAMI.2019.2914054} {The gap of
  semantic parsing: A survey on automatic math word problem solvers}.
\newblock \emph{IEEE Transactions on Pattern Analysis and Machine
  Intelligence}, 42(9):2287--2305.

\bibitem[{Zhang et~al.(2020{\natexlab{b}})Zhang, Wang, Lee, Bin, Wang, Shao,
  and Lim}]{zhang-etal-2020-graph-tree}
Jipeng Zhang, Lei Wang, Roy Ka-Wei Lee, Yi~Bin, Yan Wang, Jie Shao, and Ee-Peng
  Lim. 2020{\natexlab{b}}.
\newblock \href {https://doi.org/10.18653/v1/2020.acl-main.362} {Graph-to-tree
  learning for solving math word problems}.
\newblock In \emph{Proceedings of the 58th Annual Meeting of the Association
  for Computational Linguistics}, pages 3928--3937, Online. Association for
  Computational Linguistics.

\bibitem[{Zhong et~al.(2017)Zhong, Xiong, and Socher}]{zhongSeq2SQL2017}
Victor Zhong, Caiming Xiong, and Richard Socher. 2017.
\newblock Seq2sql: Generating structured queries from natural language using
  reinforcement learning.
\newblock \emph{CoRR}, abs/1709.00103.

\bibitem[{Zhu et~al.(2021)Zhu, Lei, Huang, Wang, Zhang, Lv, Feng, and
  Chua}]{zhu-etal-2021-tat}
Fengbin Zhu, Wenqiang Lei, Youcheng Huang, Chao Wang, Shuo Zhang, Jiancheng Lv,
  Fuli Feng, and Tat-Seng Chua. 2021.
\newblock \href {https://doi.org/10.18653/v1/2021.acl-long.254} {{TAT}-{QA}: A
  question answering benchmark on a hybrid of tabular and textual content in
  finance}.
\newblock In \emph{Proceedings of the 59th Annual Meeting of the Association
  for Computational Linguistics and the 11th International Joint Conference on
  Natural Language Processing (Volume 1: Long Papers)}, pages 3277--3287,
  Online. Association for Computational Linguistics.

\end{thebibliography}
